\documentclass{article}

\usepackage[preprint]{corl_2023} %

\title{IndoorSim-to-OutdoorReal: Learning to Navigate Outdoors without any Outdoor Experience}
\usepackage{amssymb}
\usepackage{booktabs}
\usepackage{makecell}
\usepackage[ruled,noend,linesnumbered]{algorithm2e}
\usepackage{titlesec}
\usepackage{wrapfig}
\usepackage{graphicx}
\usepackage{array}
\usepackage{multirow}
\usepackage{adjustbox}
\usepackage{colortbl}
\usepackage{tabularx}
\usepackage{nicefrac}
\usepackage{macros/mysymbols}

\usepackage{grffile}
\usepackage{xspace}
\usepackage{needspace}

\usepackage{comment} 
\usepackage{url}
\usepackage{paralist}

\usepackage[belowskip=0pt,aboveskip=5pt,font=small]{caption}
\usepackage[belowskip=0pt,aboveskip=0pt,font=small]{subcaption}
\usepackage{xcolor}
\usepackage{xfrac}
\usepackage{amsmath}
\usepackage{dsfont}
\usepackage{subcaption}
\newcommand{\xhdr}[1]{\vspace{2pt}\noindent\textbf{#1}}

\newcommand{\tableTitle}[1]{\normalsize{#1}}%
\newlength{\figwidth}%

\newcolumntype{Y}{>{\centering\arraybackslash}X}
\newcolumntype{P}[1]{\begin{center}>{\arraybackslash}p{#1}\end{center}}

\author{
  Joanne Truong$^{1,2}$, April Zitkovich$^{1}$, Sonia Chernova$^{2}$, \\
  \textbf{Dhruv Batra$^{2,3}$, Tingnan Zhang$^{1}$, Jie Tan$^{1}$, Wenhao Yu$^{1}$}\\
  $^{1}$Robotics at Google, $^{2}$Georgia Institute of Technology, $^{3}$Meta AI  \\
}

\begin{document}

\makeatletter
\let\@oldmaketitle\@maketitle%
\renewcommand{\@maketitle}{\@oldmaketitle%
\centering
\vspace{-0.2in}
\includegraphics[width=\textwidth]{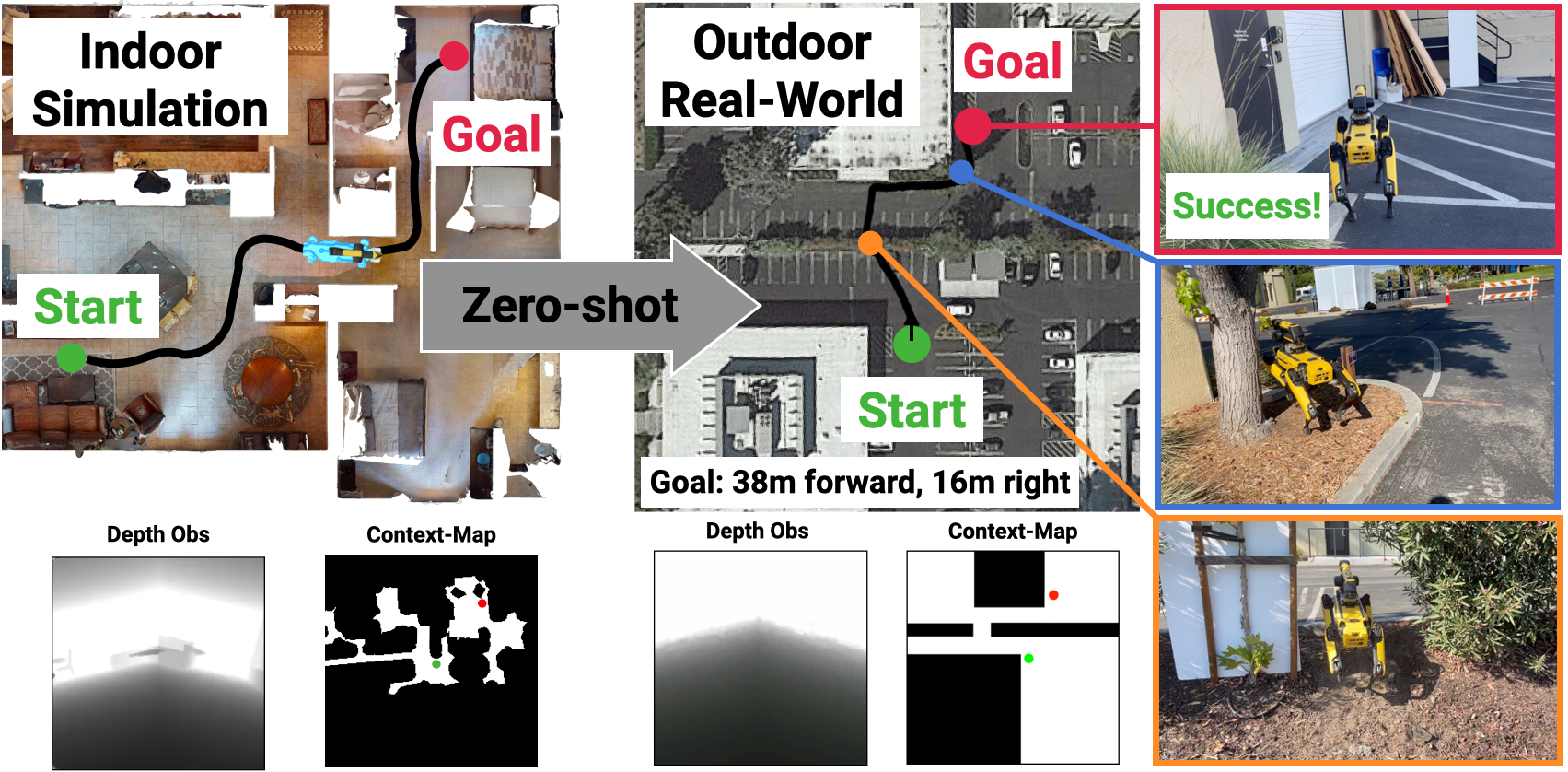}
\captionof{figure}{\small \textbf{Left:} We learn a visual navigation policy using large-scale reinforcement learning in simulated indoor environments. \textbf{Right:} 
We demonstrate zero-shot transfer of this policy for long-range outdoor navigation on the Spot robot. One key ingredient is to combine information from an inaccurate high-level map (titled Context-Map) and accurate-but-limited onboard sensing. The Context-Map is available trivially in indoor-sim and available from satellite imagery or human sketching in outdoor-real experiments. Notice how approximate and incorrect the map is (none of the cars or sidewalks show up on the map), but it does provide a crucial hint to the robot about an opening between the bushes. The robot leverages this hint to navigate up the slope to the opening in the bushes (orange), takes a shortcut between a tree and a wooden pole (blue) that is not marked on the map but visible in egocentric images, to successfully reach the goal position at the entrance of a distant building (red), utilizing the best of an outdated map and up-to-date (but partial) onboard sensing.}
\label{fig:teaser}
\vspace{-0.15in}
\bigskip}
\makeatother
\maketitle
\begin{abstract}
    We present IndoorSim-to-OutdoorReal (I2O), an end-to-end learned visual navigation approach, trained solely in simulated short-range indoor environments, and demonstrates zero-shot sim-to-real transfer to the outdoors for long-range navigation on the Spot robot. Our method uses zero real-world experience (indoor or outdoor), and requires the simulator to model no predominantly-outdoor phenomenon (sloped grounds, sidewalks, etc). The key to I2O transfer is in providing the robot with additional context of the environment (\ie a satellite map, a rough sketch of a map by a human, etc.) to guide the robot's navigation in the real-world. The provided context-maps do not need to be accurate or complete--  real-world obstacles (\eg trees, bushes, pedestrians, etc.) are not drawn on the map, and openings are not aligned with where they are in the real-world. Crucially, these inaccurate context-maps provide a hint to the robot about a route to take to the goal. We find that our method that leverages Context-Maps is able to successfully navigate hundreds of meters in novel environments, avoiding novel obstacles on its path, to a distant goal without a single collision or human intervention. In comparison, policies without the additional context fail completely.
    Lastly, we test the robustness of the Context-Map policy by adding varying degrees of noise to the map in simulation. We find that the Context-Map policy is surprisingly robust to noise in the provided context-map. In the presence of significantly inaccurate maps (corrupted with 50\% noise, or entirely blank maps), the policy gracefully regresses to the behavior of a policy with no context. Videos are available on our \href{https://www.joannetruong.com/projects/i2o.html}{project website}. 
\end{abstract}

\section{Introduction}

Much of Earth's landmass outdoors is occupied by challenging terrain that is inaccessible to wheeled robots. On the other hand, humans and legged animals are able to explore most of this landmass by finding stable footholds to navigate through these challenging terrains. 
Legged platforms provide robust locomotion, and have demonstrated successful sim2real transfer in challenging and diverse terrain such as the outdoors \cite{locomotion_terrain, locomotion_wild, kumar2021rma, agarwal2022legged}. Alongside these advancements in locomotion for legged robots, recent works in deep reinforcement learning have demonstrated success in training virtual robots to navigate efficiently in simulation before transferring the learned skills to real-world environments \cite{habitatsim2real20ral, truong2020learning, sct21iros, Partsey_2022_CVPR, truong2022kin2dyn, deitke2020robothor, Deitke2022Phone2ProcBR, chaplot2020learning, bansal2020combining}, on both wheeled and legged robots. These advances were made possible due to the development of fast, scalable simulators \cite{habitat19iccv, szot2021habitat, shen2021igibson, isaac, ai2thor, deitke2020robothor, gan2020threedworld, xiang2020sapien, todorov2012mujoco, freeman2021brax, coumans2016pybullet} and the availability of large-scale datasets of photorealistic 3D scans of indoor environments \cite{ramakrishnan2021habitat, chang2017matterport3d, chang2015shapenet}. In this paper, we seek to bring the same advancements made in indoor visual navigation to the outdoors for legged robots. 

Outdoor navigation with legged robots is different from indoor navigation, and is relatively under-studied. While indoor navigation typically span tens of meters, outdoor navigation navigating hundreds of meters to travel from one building to another. At this scale of navigation, taking the wrong turn or backtracking can be costly, and simple exploration based methods can be inefficient.

We present IndoorSim-to-OutdoorReal (I2O), which enables a quadrupedal robot to successfully navigate hundreds of meters in novel outdoor environments, around previously unseen outdoor obstacles (trees, bushes, buildings, pedestrians, etc.), in different weather conditions (sunny, overcast, sunset) -- despite being trained \textit{solely in simulated indoor environments}. The robot has never seen any outdoor environments, and has only been trained using short-range trajectories (on average 8m). We show that by training on large-scale diverse photorealistic indoor environments, we enable a robot to learn navigational skills that are transferable to the outdoors. 

\looseness=-1
We find that the key to enabling I2O is to provide the robot with additional context of its environment (\ie via a satellite image, a rough sketch by a human, etc.), which allows the learned policy to bias its search, obviating the need for costly exhaustive search and backtracking in the real-world. 
We provide additional context to the robot through a rough human sketch over a satellite image from the environment to guide the robot's navigation. These context-maps do not need to be accurate, but serve as a hint in the general directions that the robot should explore. These sketches enable a human user to specify obscure paths that may be difficult for the robot to find otherwise, or paths that are not visible in satellite images (\ie a small opening through bushes shown in Figure \ref{fig:teaser}). It is difficult to apply traditional path planning methods on these inaccurate, incomplete, and outdated context-maps. Instead, the robot must learn to use them in conjunction with observations from its onboard cameras to adapt to on-the-ground reality -- avoiding obstacles not drawn on the map (bushes, chairs, people). 

In our experiments, we find that the robot with the Context-Map is able to navigate hundreds of meters to a distant goal without a single collision or human intervention. In contrast, without the additional context, the robot completely fails at navigating long-ranges outdoors, and is only able to make minor progress towards the goal before getting stuck around obstacles. We conduct a comprehensive quantitative analysis in simulation, and demonstrate that in indoor environments, the additional context can improve success rate by 17\%, and improve path efficiency by 22\%. Additionally, we find that the Context-Map policy is surprisingly robust to noise in the provided context-map. When the maps are inaccurate (corrupted with 50\% noise, or an entirely blank map), the performance of the policy gracefully regresses back to the performance of policies trained without any context.

\section{Related Work}
\xhdr{Outdoor Navigation.} Outdoor navigation has a long history in robotics for wheeled and legged robots, and autonomous vehicles. Classical navigation approaches decompose the problem into a sense-plan-act pipeline \cite{murphy2019introduction}: a map of the environment is built, the robot is localized within the map, and a planning algorithm is used to generate a path to the goal \cite{durrant2006simultaneous, thrun2002probabilistic, fuentes2015visual}. While this classical pipeline has demonstrated successful long-range outdoor navigation \cite{morales2009autonomous,kummerle2015autonomous}, these approaches are reliant on accurate, pre-computed maps of the environment built using expensive LiDAR sensors, and are not adaptable to changes in the environment such as dynamic obstacles, or differences in weather. Our work removes the assumption of an accurate map of the environment. We use approximate and inaccurate maps to provide a hint to the robot, which are not accurate enough to support planning but contain enough information for a learned policy to improve performance using onboard sensing to adapt to the changes in the environment. 

A variety of learning-based methods have been proposed for vision-based outdoor obstacle avoidance and navigation. One prominent method is to leverage offline-learning using large real-world datasets. Muller et al. \cite{muller2005off} collect a large dataset using teleoperation, and trained a system end-to-end to map short-range observations to steering wheel angles for outdoor obstacle avoidance. Hadsell et al. \cite{hadsell2009learning} use a offline dataset of stereo images to train a long-range vision traversability classifier, which is used build a hyperbolic polar map. The map is used with traditional path planning techniques for navigation. 
Sermanet et al. \cite{sermanet2009multirange} accomplish long-range, off-road robot navigation by combining \cite{muller2005off, hadsell2009learning} to plan at at multiple ranges. Recent works investigated learning navigational affordances for long-range navigation from egocentric images by leveraging real-world datasets \cite{kahn2021badgr, kahn2021land, shah2021rapid}. The large, diverse, offline dataset consists of real-world navigation trajectories collected over months in a wide variety of environments (with the same robot embodiment) using a mix of teleoperation and random walk. Some works couple this learned module with planning on top of a pre-computed topological map \cite{shah2021ving, shah2022gnm}. In contrast, our approach does not require any explicit topological map construction, and the robot can navigate immediately in a new environment without any physical pre-exploration.
Shah et al. \cite{shah2022viking} is the most similar method to ours. They propose to use geographical hints for the task of image goal navigation outdoors. However, their work uses a large offline dataset to reason about traversability, and propose waypoints to reach long horizon goals. In contrast, our work is learned entirely in simulation and transfers to the real-world zero-shot; no real-world experience or outdoor data required and our approach is trivially adaptable to new robots because the new robot embodiment can be easily simulated.

\looseness=-1
An alternative to using real-world datasets is to leverage (outdoor) simulators. Sorokin et al. \cite{sorokin2022learning} use CARLA \cite{dosovitskiy2017carla}, an autonomous driving simulator, to train a quadrupedal robot to navigate on sidewalks by following waypoints generated by public map services (\eg Google Maps). In our work, we remove the assumption of being provided pre-computed waypoints; instead we learn to extract navigational hints directly from the context-map and onboard sensing. This enables our robot to navigate using a high-level hint, and take shortcuts through rough terrains and openings when possible. 

\xhdr{Indoor Visual Navigation \& Sim2real Transfer.} Many works in indoor visual navigation leverage large-scale learning with photorealistic simulators, such as Habitat \cite{habitat19iccv, szot2021habitat}, iGibson \cite{xia2018gibson, shen2021igibson}, or AI2-THOR \cite{ai2thor}, and techniques for reducing the sim2real gap \cite{zhu2017model, tan2018sim, tobin2017domain, peng2018sim, xie2021dynamics, truong2021bi, bousmalis2018using} to enable zero-shot sim2real transfer on wheeled and legged robots \cite{habitatsim2real20ral, truong2020learning, sct21iros, Partsey_2022_CVPR, truong2022kin2dyn, deitke2020robothor, Deitke2022Phone2ProcBR, chaplot2020learning, bansal2020combining}. 
We build upon these advances in indoor visual navigation to demonstrate long-range navigation outdoors by leveraging large-scale learning in short-range simulated indoor environments and outdated maps of the environment.

\section{IndoorSim-to-OutdoorReal Transfer}

\vspace{-0.2cm}
\subsection{Experimental Setup}
\vspace{-0.2cm}
\xhdr{Task: PointGoal Navigation}. In PointGoal Navigation (PointNav) \cite{anderson2018evaluation}, a robot is initialized in a previously unseen environment and needs to navigate to a goal location (\ie \textit{``go to} $\Delta$x, $\Delta$y"). The robot has access to egocentric RGB-D observations, and an egomotion sensor which is used to calculate the goal location relative to the robot's current pose. The robot operates within constraints of maximum number of steps per episode (500 steps per 50m from the goal location) and velocity limits ($\pm$0.5 $\sfrac{m}{s}$ for linear and $\pm$0.3 $\sfrac{rad}{s}$ for angular velocities). An episode is considered successful if the robot reaches within 0.425m (half the width of the Spot robot) of the goal location. For evaluation, we report the robot's success rate (SR), and Success inversely weighted by Path Length (SPL) \cite{anderson2018evaluation}, which measures the robot's path efficiency. 

\begin{wrapfigure}{r}{0.5\textwidth}
\vspace{-0.7cm}
  \begin{center}
    \includegraphics[width=0.5\textwidth]{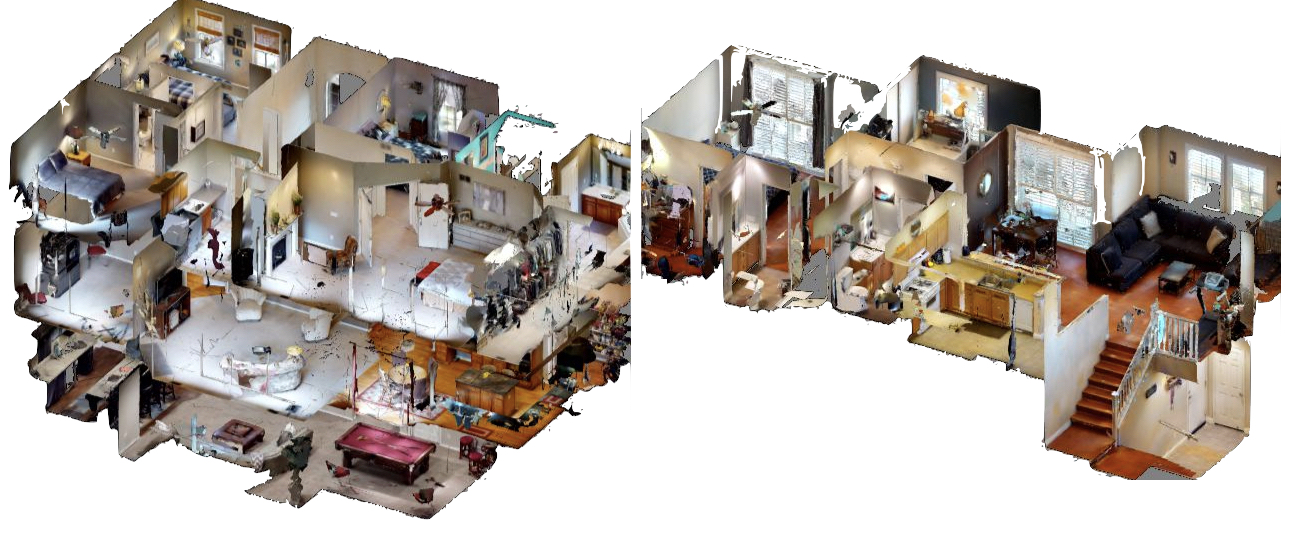}
    \caption{Examples of indoor environments used for training and evaluation in simulation.}
    \vspace{-0.9cm}
   \label{fig:hm3d}
   \end{center}
\end{wrapfigure}
\xhdr{Dataset.} We use the Habitat-Matterport (HM3D) \cite{ramakrishnan2021habitat} and Gibson \cite{xia2018gibson} 3D datasets, which consists of over 1000 scans of real-world indoor environments (homes, offices, etc.) consisting of realistic clutter (tables, chairs, etc.). We use the training and evaluation navigation episodes from \cite{truong2022kin2dyn}, which were generated to result in complex paths (up to 30m) that are navigable by the Spot robot.

\subsection{Context-Guided PointGoal Navigation}
\begin{wrapfigure}{r}{0.5\textwidth}
\vspace{-0.7cm}
  \begin{center}
    \includegraphics[width=0.5\textwidth]{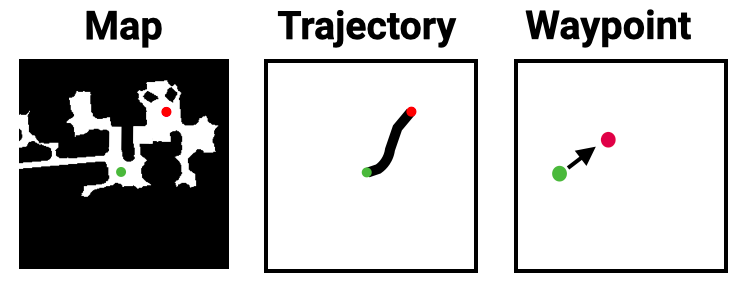}
    \caption{Three kinds of contextual input.}
    \vspace{-0.4cm}
   \label{fig:context_inputs}
   \end{center}
\end{wrapfigure}
We aim to leverage additional context information freely accessible through Google Maps satellite images for the task of long-range PointNav outdoors. 
We denote this variant of PointNav with additional environmental context, as Context-Guided PointNav (ContextNav). 
The additional context input (Figure \ref{fig:context_inputs}) can be in the form of an outdated map (Context-Map), a ground truth trajectory (Context-Trajectory), or ground truth waypoints for the robot to follow (Context-Waypoint). We train and compare the performance of all three forms of contextual input in simulation, with Context-Trajectory and Context-Waypoint policies serving as the upperbound for performance when provided with oracle context. However, we find in our simulation experiments that that Context-Trajectory and Context-Waypoint policies are susceptible to noise in the contextual input, while Context-Map policies are robust to noise. This points to Context-Maps being the best context modality for sim2real transfer. In the real-world, we present results using Context-Maps. Additional details are described in Section \ref{sec:noisy-context-vars} in the Appendix.

\begin{wrapfigure}{r}{0.5\textwidth}
\vspace{-0.5cm}
  \begin{center}
    \includegraphics[width=0.5\textwidth]{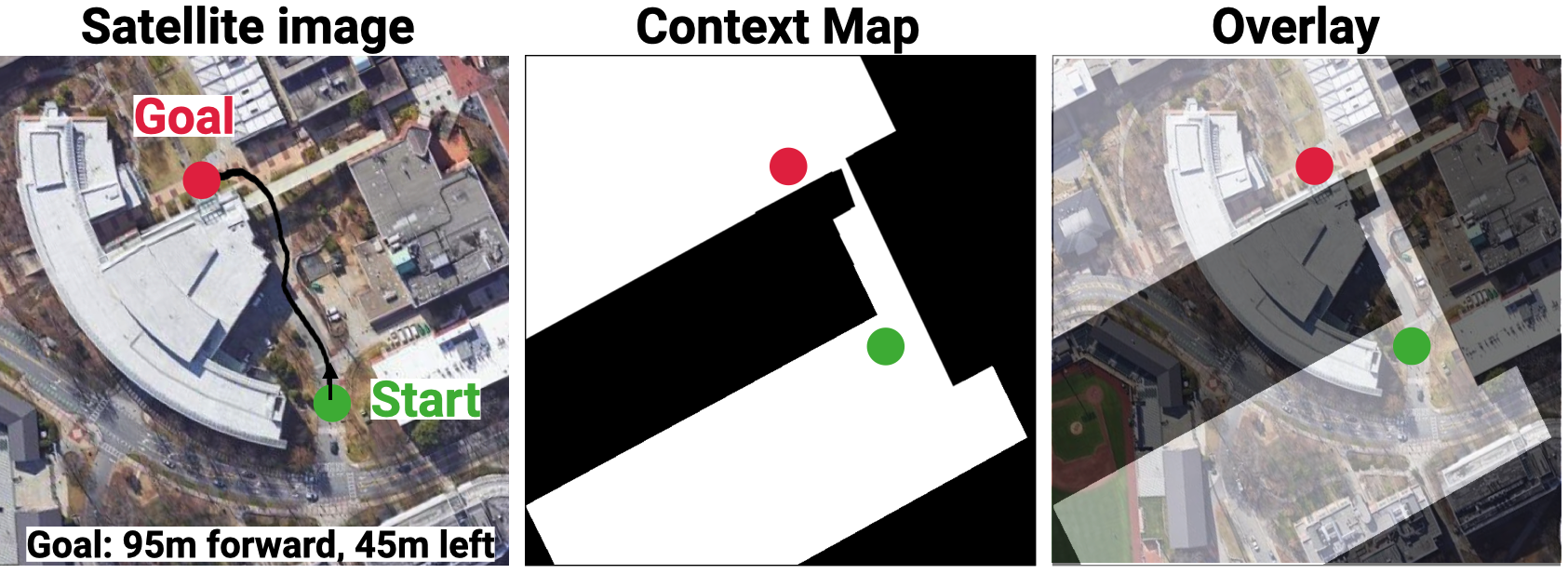}\\
    \includegraphics[width=0.5\textwidth]{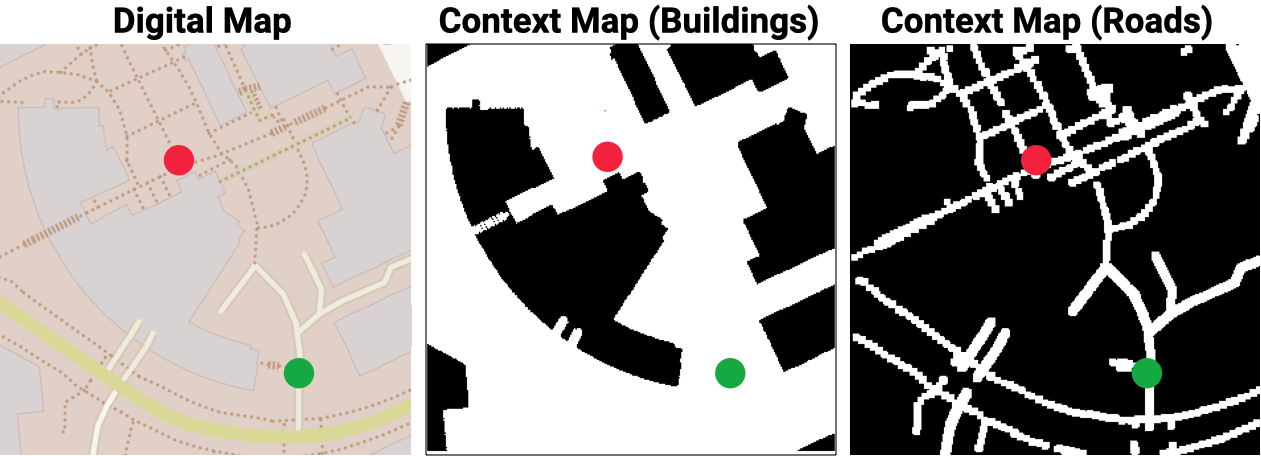}\\
    \caption{\textbf{Top Row: }Using a satellite image of the area from Google Maps(left), a human operator sketches a rough map to serve as context input to the robot (middle). Notice how defective the map is (right): large parts or entire building are missing, no roads or sidewalks are shown; but crucially, the map contains a hint for an opening to get to the goal. Context-Maps for real-world routes are shown in Section \ref{sec:context-maps-real} of the Appendix. \textbf{Bottom Row:} Context-Maps can also be automatically generated. Starting from the digital map (left), we automatically generate context-maps by segmenting out buildings (middle), or roads (right).}
    \vspace{-0.5cm}
   \label{fig:map_overlay}
   \end{center}
\end{wrapfigure}

The Context-Map input does not need to be very accurate, but should serve as a rough guide for general directions that the robot should explore (\ie a satellite map showing roughly where buildings are). Consequently, the robot must use observations from its camera to adapt to novel obstacles or clutter present in the environment but absent on the map, and be willing to take shortcuts available in the world but shown as obstacles on the map. We represent the map as a top-down occupancy map, which can be obtained in the real-world by converting a satellite image to illustrate occupied and freespace through a human sketch, or through an automated process. 
Using a binary occupancy map provides a few benefits over directly using satellite images. An abstracted binary map allows the maps to be used for both indoor and outdoor navigation, while satellite images are only applicable to outdoor navigation. Additionally, by using human sketches for context maps, the human operator can give hints to the robot about paths that may not be easily visible in the satellite image, or from the robot's initial position (\ie openings in bushes). In our experiments, we use human-sketched Context-Maps, however these maps can easily be generated automatically by postprocessing a digital map. Figure \ref{fig:map_overlay} shows an example of a human-sketched Context-Map (top row), and an automatically generated Context-Map (bottom row). Additional details on automatically generated Context-Maps are described in Section \ref{sec:auto-maps} in the Appendix.

\begin{figure*}[t]
  \centering%
  \includegraphics[width=\textwidth]{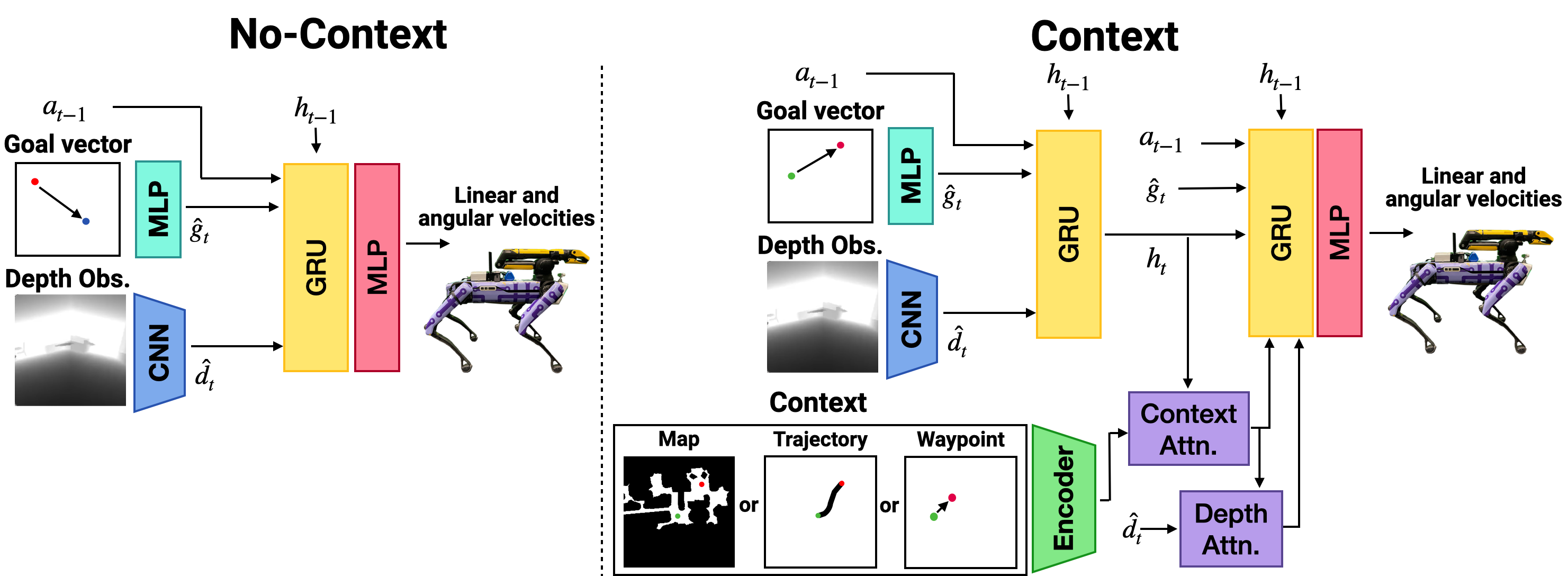}
  \caption{\textbf{Left:} The No-Context PointNav policy architecture consists of a 3-layer CNN to process depth images from the robot's camera, and a MLP to process the goal vector. The policy is a GRU, and outputs linear and angular velocities for the Spot robot to follow. \textbf{Right:} The Context-Guided PointNav architecture includes an additional encoder to process the context-input (top-down map or waypoint). We compute the scaled dot product attention between the map and the depth image, and use a second GRU to process the attended features.}
  \label{fig:architecture}
\end{figure*}

\xhdr{Policy Architecture.} We train a high-level visual navigation policy entirely in simulation using deep reinforcement learning. We first describe the architecture of a PointNav policy that operates without the additional context (No-Context) from \cite{truong2022kin2dyn}, then we outline our extension to the policy architecture to incorporate additional environmental context. The policy takes as input an egocentric depth image, a goal vector, and the action at the previous timestep. The goal vector is represented as the distance and heading relative to the robot's current pose. The output of the policy is the desired center-of-mass linear and angular velocities ($v_x$, $\omega$) for the robot to follow. The depth observations are processed using a 3 layer CNN visual encoder, and the goal vector is encoded using a linear layer. These features are fed into a 1 layer GRU, followed by a single linear layer which parameterizes a Gaussian action distribution from which the action is sampled. The policy architecture is shown in Figure \ref{fig:architecture}, left. Additional details are described in Section \ref{sec:no-context-arch} of the Appendix.

Next, we describe how to incorporate additional context information for ContextNav. We first describe map-based context inputs (Context-Map and Context-Trajectory), followed by waypoint-based context inputs (Context-Waypoint). In simulation, we use freely accessible top-down maps of indoor environments as additional context to aid navigation. The context-map is represented as a $2 \times N \times N$ matrix, where $N \times N$ represents the size of the map (we use $N = 100$). For Context-Map, the first channel of the map is an allocentric occupancy map, in which the cells denote obstacles (0) or freespace (1). For Context-Trajectory, the first channel of the map is an allocentric shortest-path trajectory map obtained by running A* on the top-down map of the environment. In the trajectory map, the cells denote the shortest path trajectory (1), and 0 otherwise. For both map-based context inputs, the allocentric map is rotated according to the robot's current heading. The second channel of the map illustrates the agent's current location in the top-down map (1), and the location of the goal coordinate with a small disk (1), and 0 otherwise. An example of the second channel is overlaid in green and red in the Context-Map in Figure \ref{fig:context_inputs}, left. We process the map context inputs using a ResNet18 visual encoder \cite{he2016resnet}. To provide waypoints as context to the navigation policy, we calculate the next closest waypoint along the shortest path to the goal within 1m from the robot's current position. The waypoint is specified using polar coordinates $(r, \theta)$, representing the distance and heading from the robot's current position. The waypoint is encoded using a two-layer MLP, with a 512 hidden dimension. We compute the scaled dot-product attention \cite{vaswani2017attention} between the depth and context features, and pass the attended features into a second GRU. Additional details are described in Section \ref{sec:context-arch} of the Appendix.

\xhdr{Training details.} We train all our policies for 500M steps using DD-PPO \cite{ddppo}, a distributed reinforcement learning method, in the Habitat simulator \cite{habitat19iccv, szot2021habitat}. The reward function is derived from \cite{truong2022kin2dyn}, which includes a dense reward for following the shortest path to the goal, a terminal reward for successful episodes, a slack penalty to incentivize the robot to reach the goal quickly, a penalty for collisions, and a penalty for backward velocities, which can lead to collisions and hurts performance. 
Additional details on the reward function are described in Section \ref{sec:reward} of the Appendix.

\subsection{Techniques to Aid IndoorSim-to-OutdoorReal Transfer.}
\label{sec:bot}
Empirically, we found that several additional key techniques were needed to enable IndoorSim-to-OutdoorReal transfer. We use these techniques for all policies tested in the real-world.

\xhdr{Indoor-to-Outdoor Transfer.} 
Two of the main differences between indoor and outdoor navigation are 1) navigation length (short-range vs. long-range), and 2) terrain type (flat vs. rocky/sloped). First, we found that traditional PointNav policies were highly sensitive to the goal vector. Since the policies were only trained in simulation, and typically see trajectories $\sim$8m away, the policies failed to generalize to longer-range goals. We normalize the goal vector by using the log of the goal distance, which enabled longer-range navigation. Next, we found that naively trained PointNav policies had difficulty navigating up slopes. Since slopes are infrequent in indoor environments, when the robot sees a slope outdoors, the slope appears as a large obstacle in the robot's depth camera, and thus the robot avoids walking up the slope. To enable the robot to walk up and down slopes, we randomize the pitch of the camera during training by $\pm 30^{\circ}$. Adding pitch randomization artificially adds slopes to the robot's observation during training, which enables the robot to walk up slopes in the real-world.

\xhdr{Sim-to-Real Transfer.} 
We use a few techniques to enable zero-shot sim2real transfer. In simulation, instead of modeling the robot's low-level locomotion control, we use kinematic control as an approximation for the robot's movement. In kinematic control, the robot is moved to its next state via Euler integration at 2Hz without running full rigid-body physics. If the robot were to collide at the next state, we simply keep the robot in place. Kinematic control was shown in \cite{truong2022kin2dyn} to lead to better sim-to-real transfer through faster simulation, as compared to dynamic control. 
\begin{wrapfigure}{r}{0.5\textwidth}
\vspace{-0.5cm}
  \begin{center}
    \includegraphics[width=0.5\textwidth]{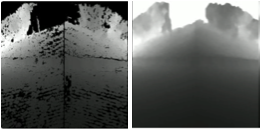}
    \caption{\textbf{Left:} Raw depth images from the Spot robot's front cameras. \textbf{Right:} We filter the depth images using depth completion from \cite{ku2018defense} to better match images from simulation.}
    \vspace{-0.4cm}
   \label{fig:depth_filtering}
   \end{center}
\end{wrapfigure}
Next, we filter the depth images from Spot's depth cameras in the real-world to better match observations from simulation. The raw (noisy) depth images from Spot are show in in Fig 6 (left). We use depth completion from \cite{ku2018defense} to fill in holes and smooth the image. Additionally, we set the pixels further away from the max depth range of the camera (3.5m) to white to match the depth images from simulation. The filtered depth image is shown in Figure \ref{fig:depth_filtering} right. 
Lastly, improve the robustness of our policies to missing pixels in real-world depth images, we add add Random Erasing noise \cite{zhong2020random} to our depth observations during training.

\subsection{Real-world Setup}
\looseness=-1
We use the Boston Dynamics (BD) Spot robot in simulation and the real-world. Our navigation policy outputs high-level velocity commands, and we rely on BD's low-level controller for movement. We use Spot's two front-facing Intel Realsense D430 depth cameras for visual inputs to our policy. We disable the BD obstacle avoidance in order to isolate the performance of our policy from confounding factors; however, a human operator was vigilant to terminate the episode (and report it as a failure) if a collision were imminent. The BD API includes two modes for high-level navigation, but both methods have limitations and cannot be used directly for long-range outdoor navigation. We detail the two modes and their failure cases for long-range navigation in Section \ref{sec:bd-api} of the Appendix.

\section{Results}
\label{sec:exp}

In this section, we address the following questions: 1) Can a robot trained to navigate solely in simulated short-range indoor environments zero-shot transfer to long-range outdoors in the real-world? 2) How vital is the use of the additional context? 3) How robust is the policy to noise?

\subsection{Zero-shot IndoorSim-to-OutdoorReal Navigation} 
\begin{figure*}[t]
  \centering%
  \includegraphics[width=\textwidth]{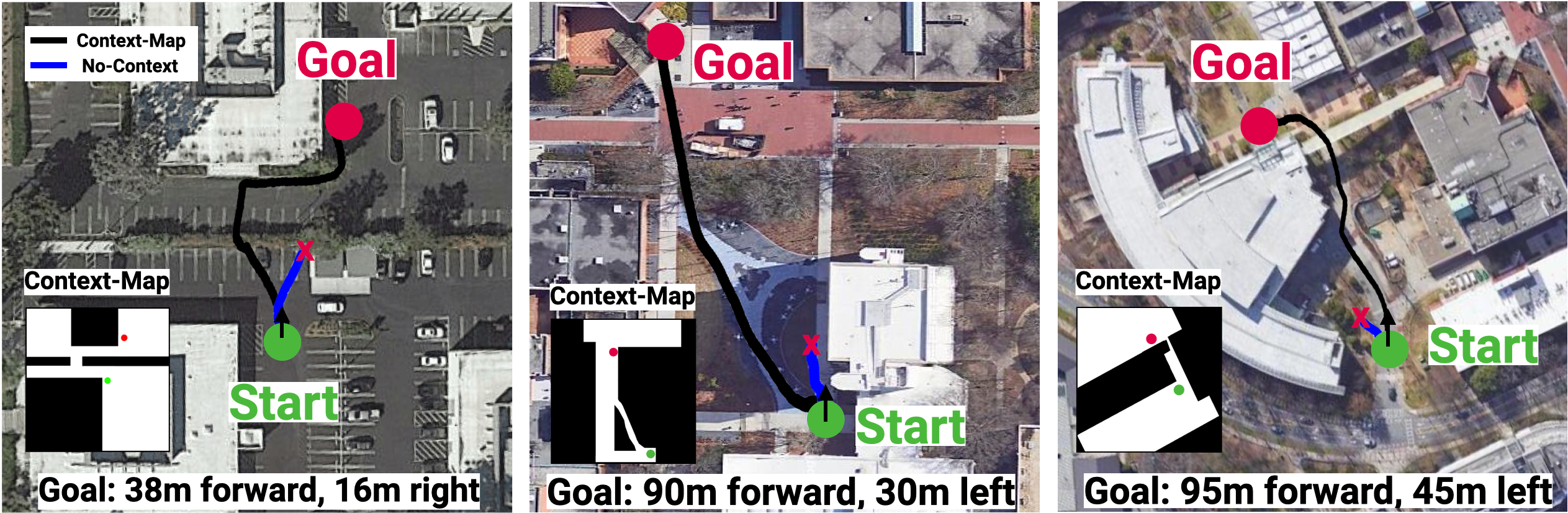}
  \caption{We test our learning-based navigation policies in novel environments for long-range outdoor navigation using the Spot robot. The three routes in the real-world contain many obstacles including bushes, buildings, cars and pedestrians that the robot has never seen before during training. We show the trajectories taken by our Context-Map policies in the outdoors (black); the policy is able to navigate hundreds of meters outdoors to successfully reach the goal 100\% of the time. In comparison, the No-Context policies (blue) try to make a direct beeline towards the goal, and get stuck on obstacles, thereby failing to reach the goal. 
  }
  \label{fig:real_world_runs}
\end{figure*}

We investigate zero-shot sim2real indoor2outdoor navigation. We task the robot with navigating to 3 long-range goals outdoors (shown in Figure \ref{fig:real_world_runs}), with many real-world obstacles present (bushes, buildings, cars, tables, pedestrians, etc.) and different weather conditions (sunny, overcast, sunset). 
We report the average success rate (SR) and distance travelled across 3 runs for each method in Table \ref{tab:real_world_res}. Videos of our experiments are available on our \href{https://www.joannetruong.com/projects/i2o.html}{project website}. 

\xhdr{No-Context Real-world Outdoor Navigation.}
First, we test to see if policies with No-Context can directly transfer to long-range navigation outdoors. 
The robot is required to navigate around large obstacles to reach the goal successfully. However, we find that the No-Context policy immediately makes a beeline towards the goal using the goal vector for guidance. 
\begin{wraptable}{r}{0.6\textwidth}
	\vspace{-0.1cm}
	\begin{center}
        \resizebox{0.6\textwidth}{!}{
		\begin{tabular}{ccccc}
			\toprule
                \multirow{2}{*}{\textbf{Route \#}} & \multirow{2}{*}{\textbf{Goal}} & \multirow{2}{*}{\textbf{Method}} & \multirow{2}{*}{\textbf{SR} $\uparrow$} & \textbf{Distance}\\
                & & & & \textbf{Travelled (m) $\uparrow$} \\
			\midrule
                \multirow{2}{*}{1} & 38m Forward, & No-Context & 0.0 & 16.6\tiny{$\pm$0.1} \\
			& 16m Right & Context-Map & \textbf{100.0} & \textbf{63.4}\tiny{$\pm$2.5} \\
                \midrule
                \multirow{2}{*}{2} & 90m Forward, & No-Context & 0.0 & 9.7\tiny{$\pm$3.4} \\
			& 30m Left & Context-Map & \textbf{100.0} & \textbf{112.2}\tiny{$\pm$1.8} \\
                \midrule
                \multirow{2}{*}{3} & 95m Forward, & No-Context & 0.0 & 5.1\tiny{$\pm$0.3} \\
			& 45m Left & Context-Map & \textbf{100.0} & \textbf{129.8}\tiny{$\pm$2.8} \\
			\bottomrule
		\end{tabular}
            }
	\end{center}
	\caption{\small{We test the No-Context and Context-Map policies on 3 long-range outdoor routes (Figure \ref{fig:real_world_runs}).}} 
	\label{tab:real_world_res}
	\vspace{-0.3cm}
\end{wraptable}

In indoor environments, there are more obstacles around the robot, such as walls, that guide the robot to avoid making a beeline to the goal. 
In the outdoors however, the robot makes a beeline presumably led by the depth sensors that indicate plenty of free-space around the robot. This leads the robot to wander into obstacles such as bushes, resulting in unsuccessful episodes. The No-Context policy completely fails to navigate in all three routes, each only making minor progress to the goal before an operator had to intervene to prevent a collision (Table \ref{tab:real_world_res}). The trajectories taken by the No-Context policy is shown in blue in Figure \ref{fig:real_world_runs}. In the first route, we task the robot with navigating from one building to another (38m forward, 16m to the right). The shortest path to the next building requires passing through a small opening to the left of the robot between two bushes that is not visible to the robot from the start. Since the goal vector indicates that the goal is to the right, the robot starts walking to the right, and gets stuck searching by the bushes, resulting in an episode failure.

\begin{wrapfigure}{r}{0.5\textwidth}
\vspace{-0.4cm}
  \centering%
  \includegraphics[width=\linewidth]{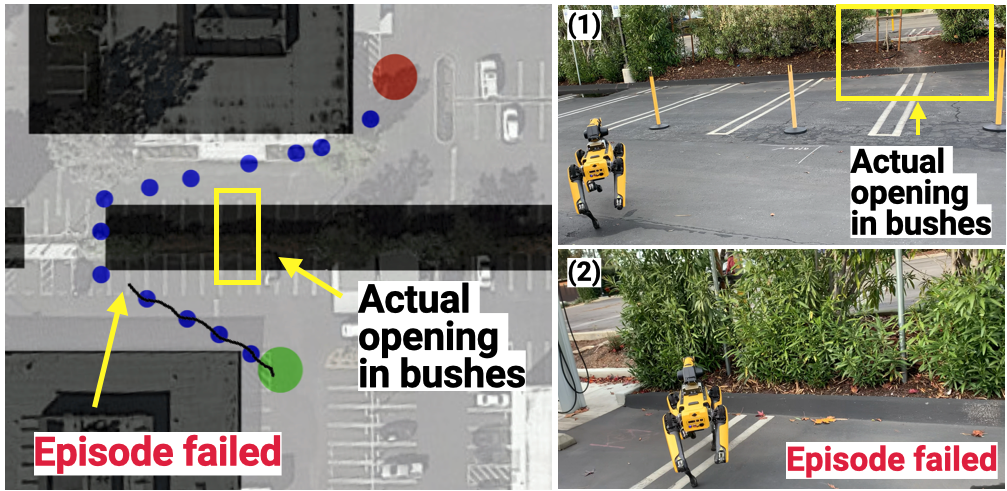}
  \caption{\textbf{Left:} We run RRT* on an outdated context-map, to find waypoints (blue) along the shortest path to the goal. We pass the waypoints into a Context-Waypoint policy to navigate to the goal (black). \textbf{Right:}  Since the map is inaccurate, the waypoints lead the robot past the opening in the bushes (1), leading to an episode failure (2).}
  \vspace{-0.2cm}
  \label{fig:rrt_baseline}
\end{wrapfigure}

\xhdr{Traditional Planning Real-world Outdoor Navigation.}
Next, we test to see if outdated context-maps can aid in navigating outdoors. We first investigate leveraging classical approaches to plan using the context-map. We run RRT* on the context map to generate a list of waypoints to the goal. These waypoints are then used with the Context-Waypoint policy, which was trained in simulation to follow waypoints along the shortest route to the goal. 
We find that the Context-Waypoint policy is able to successfully navigate to each successive waypoint. However, the robot ended up missing the actual opening in bushes because the waypoints were generated from an outdated map (shown in Figure \ref{fig:rrt_baseline}). This confirms our conjecture that classical planning based approaches are highly sensitive to the map input, and are not able to adapt to gross inaccuracies present in the maps. In the real-world, creating a perfect, complete, and always-up-to-date map is not realistic. We show in the next section that the Context-Map policy can successfully reach the goal despite being given the same outdated, inaccurate map.

\looseness=-1
\xhdr{Context-Guided Real-world Outdoor Navigation.}
Finally, we test our Context-Map policy outdoors. We provide the robot with rough sketches indicating preferred paths for the robot to take for each route (Figure \ref{fig:real_world_runs}). Notice how rough and incomplete the provided maps are-- obstacles such as cars, trees, or chairs are not shown on the map, and only rough hints for an opening to the goal is depicted. While classical planning methods that used these \textit{exact same} outdated maps failed to reach the goal, we find that the Context-Map policy is able to leverage the context-maps to bias its search during outdoor navigation, and successfully reach the distant goal location 100\% of the time (Table \ref{tab:real_world_res}), without a single collision or human intervention. Shah et al. \cite{shah2022viking} report similar results using 42 hours of real-world robot navigation data. In contrast, our approach uses zero real-world experience.

We find that our policies are able to maintain a balance between the Context-Maps, and what it sees in its visual inputs. The robot leverages depth images to avoid clutter/dynamic obstacles, and Context-Maps are used for high-level guidance. While Context-Maps provide a means for the operator to specify a preferred route for the robot to take, the robot may take shortcuts along the way that are not present in the map when its vision senses freespace. However, we do not observe the robot finding a drastically `better' solution than what is hinted on the map (\ie, a better path to the right, when the map indicates free space only on the left), due to the robot’s limited vision (recall that the depth cameras have a range of only $\sim$3m).
With our approach, the robot was able to navigate around dynamic obstacles such as pedestrians and real-world clutter despite these obstacles not being drawn directly on the context-map. Our approach is also not limited to 2D navigation; the robot navigates up slopes (Route 1), and can navigate in various terrains including dirt slopes and grass.

\subsection{Simulation Results}
\xhdr{Indoor Navigation.} 
We conduct extensive experiments in simulation to systematically quantify the value of using the additional context input. 
We first evaluate using 1200 episodes of indoor environments from the HM3D and Gibson scenes \cite{truong2022kin2dyn}. We report the average success rate (SR) and SPL across 3 seeds in Table \ref{tab:hm3d_val}. 

\begin{wraptable}{r}{0.5\textwidth}
	\vspace{-1.0cm}
	\begin{center}
        \resizebox{0.5\textwidth}{!}{
		\begin{tabular}{cccccc}
			\toprule
                \textbf{\#} & \textbf{Method} & \textbf{Context} & \textbf{SR $\uparrow$} & \textbf{SPL $\uparrow$} \\
			\midrule
			1 & No-Context & - & 78.7\tiny{$\pm$8.2
			} & 56.4\tiny{$\pm$4.4}  \\
			2 & Context & Map &  95.4\tiny{$\pm$0.2} & 81.0\tiny{$\pm$2.4} \\
                \midrule
			3 & Context & Waypoint &  96.4\tiny{$\pm$0.4} & 86.3\tiny{$\pm$0.6} \\
			4 & Context & Trajectory &  96.5\tiny{$\pm$0.1} & 84.5\tiny{$\pm$1.2} \\
			\bottomrule
		\end{tabular}
            }
	\end{center}
	\caption{The Context-Map policy navigates just as well as policies that are provided oracle context (waypoint, trajectory). Additionally, we see that the Context-Map policy achieves a 16.7\% higher success rate and 24.6\% higher SPL than the No-Context policy.}
	\label{tab:hm3d_val}
	\vspace{-0.3cm}
\end{wraptable}
To get an upperbound for performance, we evaluate 2 policies that get different levels of oracle context-- a policy that receives as input the next waypoint along the shortest path to the goal (row 3 in Table \ref{tab:hm3d_val}), and a policy that receives a top-down map with the shortest path trajectory drawn on the map (row 4 in Table \ref{tab:hm3d_val}). We find that the Context-Map policies achieve a similar high success rate and SPL as the policies that get as input oracle context (95.4 vs. 96.5 SR, and 81.0 vs. 86.3 SPL; Table \ref{tab:hm3d_val} rows 2-4), demonstrating that the policy is able to utilize its given context-map to navigate to the goal. In contrast, the No-Context policies achieve a lower success rate of 78.7\% (-16.7\% SR, rows 1 and 2 in Table \ref{tab:hm3d_val}), demonstrating that the use of the additional context is useful in guiding the robot in its navigation to successfully reach the goal. We also see a +24.6\% SPL between the two policies, demonstrating that the robot can navigate to the goal more efficiently using the added context. These experiments in demonstrate that Context-Map policies can navigate through obstacles in cluttered environments, and is capable of complex navigating through multiple rooms.

\begin{figure}[h]
  \centering%
  \includegraphics[width=\textwidth]{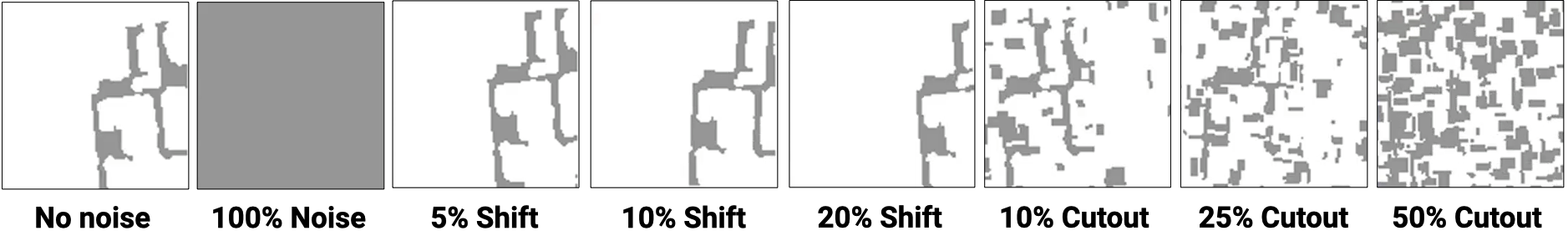}
  \caption{We use shift and cutout noise to degrade the top-down map obtained in simulation.}
  \label{fig:noisy_maps}
\end{figure}

\xhdr{Indoor Navigation using Outdated Maps.} 
\label{sec:outdated-maps}
We test the robustness of our policy by adding varying degrees of noise to the context-map, shown in Figure \ref{fig:noisy_maps}. Specifically, we use: 
\begin{asparaenum}
\item Shift noise, which randomly shifts the map provided to the robot in any direction. This experiment simulates localization errors that may occur on the robot.
\item Cutout noise, which randomly adds patches of free space or obstacle space to the map. This experiment simulates giving the robot a outdated map (\ie a map with additional or missing obstacles in the environment). 
\end{asparaenum}

We find that our approach, trained entirely using perfect maps, is surprisingly robust to the noisy maps.
At 50\% cutout noise, the original top-down map is barely visible (Figure \ref{fig:noisy_maps}), and planning algorithms would be unable to find any feasible path to the goal. 
In contrast, our policy is still able to navigate to the goal successfully 81.4\% of the time (row 7 in Table \ref{tab:noisy_hm3d_val}).

\begin{wraptable}{r}{0.45\textwidth}
  \vspace{-0.3cm}
    \begin{center}
        \resizebox{0.45\textwidth}{!}{
        \begin{tabular}{ccccc}
            \toprule
                \multirow{2}{*}{\textbf{\#}} & \multicolumn{2}{c}{\textbf{Eval Noise}} & \multirow{2}{*}{\textbf{SR $\uparrow$}} & \multirow{2}{*}{\textbf{SPL $\uparrow$}} \\
                \cmidrule(l{4pt}r{4pt}){2-3}
                & \textbf{Type} & \textbf{Percent} & & \\
            \midrule
               1 & - & - & 95.4 \tiny{$\pm$0.2} & 81.0 \tiny{$\pm$2.4} \\
                \midrule
               2 & \multirow{3}{*}{Shift} & 5\% & 93.0 \tiny{$\pm$0.9} & 75.1 \tiny{$\pm$2.4} \\
               3 & & 10\% & 84.0 \tiny{$\pm$0.5} & 62.5 \tiny{$\pm$2.3} \\
               4 & & 20\% & 73.2 \tiny{$\pm$1.6} & 51.6 \tiny{$\pm$1.1} \\
               \midrule
                5 & \multirow{3}{*}{Cutout} & 10\% & 91.5 \tiny{$\pm$0.2} & 72.3 \tiny{$\pm$2.2} \\
                6 & & 25\% & 87.6 \tiny{$\pm$0.8} & 67.4 \tiny{$\pm$2.3} \\
                7 & & 50\% & 81.4 \tiny{$\pm$2.2} & 60.0 \tiny{$\pm$1.0} \\
               \midrule
               8 & & 100\% & 79.8 \tiny{$\pm$1.2} & 59.8 \tiny{$\pm$0.9} \\
            \bottomrule
        \end{tabular}
            }
    \end{center}
    \caption{We evaluate Context-Map policies with varying degrees of noise to the map.}
    \label{tab:noisy_hm3d_val}
  \vspace{-1.1cm}
    \end{wraptable}
When the Context-Map policy is given a map that is completely freespace (100\% noise, row 8 in Table \ref{tab:noisy_hm3d_val}) the policy behavior regresses to the No-Context policy (79.8 vs. 78.7 SR, and 59.8 vs. 56.4 SPL). Thus, our approach exhibit the best of both worlds -- utilizing the information in the map when it's available but never underperforming a map-free approach. For completeness, we additionally train policies explicitly on noisy maps and report the results in Section \ref{sec:noisy-maps} of the Appendix. We find that by training with noisy maps in simulation, the policy performance is even more robust when evaluated with noisy maps.

\section{Discussion \& Limitations} Our results provide compelling evidence for rejecting the (admittedly reasonable) hypothesis that a new simulator must be designed for every new scenario we wish to study. 
This is especially important for environments that are challenging to design in simulation, such as the outdoors. Traditional approaches for `virtualizing' a real-world environment to import into a simulator leverage infrared projection based 3D scanners. However, in the outdoors, infrared light interference from the sun prohibits scanning these environments accurately. Instead, our approach shows that simulators can be used for zero-shot real-world transfer without having to design and model the deployment scenario apriori. Our work uses zero real-world experience (indoor or outdoor) and requires the simulator to model no predominantly-outdoor phenomenon (terrain, ditches, sidewalks, cars, etc). The power in the approach comes from a combination of a robust locomotion system on Spot (which itself uses a well-designed classical robotics stack), low sim-vs-real gap in depth and map sensors, and the power of large-scale learning. 
Similar to literature in cognitive science \cite{foo2005humans} that find that humans use approximate cognitive maps (\ie through landmarks) to guide navigation, our results indicate that providing robots with approximate, high-level maps can enable long-range navigation in novel outdoor environments. 

Our approach suffers from a few limitations. First, there may still be certain instances in which the robot may fail such as when both the depth observation and Context-Map may suggest freespace in front of the robot despite an obstacle being present (narrow railings, glass walls, etc.). To avoid such failures, it is important to incorporate other sensing modalities such as RGB images or proprioception input \cite{fu2021coupling}, or to enable the robot with additional actions (look up, look down, etc.) to increase the robot's sensing. Additionally, our approach relies on the BD API for localization, which may drift over time as the robot navigates even longer distances. In the future, we would like to remove the assumption accurate localization sensors on the robot.

While our approach demonstrates that large-scale learning with simulated data alone can enable outdoor navigation, prior works have also demonstrated impressive results using real-world data. In the future we would like to explore combining learning with both real-world and simulated data to further improve our policy. Training the policy from scratch in the physical outdoor environment is intractable and may cause damage to the robot over time, but it may be beneficial to fine-tune a pre-trained policy in the real-world. An alternative is to leverage offline trajectories from the real-world to augment the simulator \cite{truong2021bi, golemo2018sim}.

\section{Conclusion} In this work, we present a framework for IndoorSim-to-OutdoorReal transfer for navigation. Our navigation policy is trained solely in simulated short-range trajectories from indoor datasets, yet is able to navigate long ranges outdoors in the real-world. The navigation policy is robust to novel environments and obstacles such as sloped grounds, bushes, and trees never seen during training. Keys to the system's success are a set of sim-to-real techniques that enable the policy to handle real outdoor environments, as well as the addition of context in the form of a rough sketch provided by a human, which guides the robot's navigation. 
This work opens up navigation research to the less explored domain of the rich and diverse outdoors environments.

\acknowledgments{
The Georgia Tech effort was supported in part by NSF, AFRL, DARPA, ONR YIPs, ARO PECASE. JT was supported by an Apple Scholars in AI/ML PhD Fellowship. The views and conclusions are those of the authors and should not be interpreted as representing the U.S. Government, or any sponsor.

\noindent \textbf{License for dataset used.} Gibson Database of Spaces. License at \url{http://svl.stanford.edu/gibson2/assets/GDS_agreement.pdf}

Satellite maps used are from Google Maps. Digital maps used are from OpenStreetMap. 
}

\bibliography{bib/main}  %

\clearpage
\section{Appendix}

The appendix is structured as follows:
\begin{enumerate}
    \item Indoor Navigation using Noisy Context Variants
    \item Improved Robustness for Indoor Navigation using Outdated Maps
    \item PointNav Policy Architecture
    \item ContextNav Policy Architecture
    \item Context-Maps in the Real-World
    \item Auto-Generated Context-Maps
    \item Boston Dynamics Navigation API
    \item Reward functions
    \item Additional Simulation Details
\end{enumerate}

\subsection{Indoor Navigation using Noisy Context Variants} 
\label{sec:noisy-context-vars}

\begin{wraptable}{r}{0.5\textwidth}
	\vspace{-0.8cm}
	\begin{center}
        \resizebox{0.5\textwidth}{!}{
		\begin{tabular}{clcccc}
			\toprule
                \textbf{\#} & \textbf{Method} & \textbf{Eval Noise} & \textbf{SR $\uparrow$} & \textbf{SPL $\uparrow$} \\
			\midrule
               1 & \multirow{4}{*}{\parbox{0.5cm}{Context-Waypoint}} & No noise & 96.4\tiny{$\pm$0.4} & 86.3\tiny{$\pm$0.6} \\
               2 & & 0.25m Shift & 88.3\tiny{$\pm$1.6} & 62.3\tiny{$\pm$3.4} \\
               3 & & 0.5m Shift & 77.3\tiny{$\pm$0.3} & 47.5\tiny{$\pm$2.8} \\
               4 & & 1.0m Shift & 36.7\tiny{$\pm$2.0} & 18.5\tiny{$\pm$1.4} \\
               \midrule
                5 & \multirow{4}{*}{\parbox{0.5cm}{Context-Trajectory}} & No noise & 96.5\tiny{$\pm$0.1} & 84.5\tiny{$\pm$1.2} \\
                6 & & 5\% Shift & 68.5\tiny{$\pm$1.4} & 42.7\tiny{$\pm$1.3} \\
                7 & & 10\% Shift & 57.8\tiny{$\pm$2.8} & 33.1\tiny{$\pm$0.1} \\
                8 & & 20\% Shift & 46.9\tiny{$\pm$1.9} & 25.7\tiny{$\pm$0.8} \\
			\bottomrule
		\end{tabular}
            }
	\end{center}
	\caption{We evaluate Context-Waypoint and Context-Trajectory policies with varying degrees of noise, and report the average across 3 seeds. We find that these methods are more susceptible to noisy inputs than Context-Map policies.}
	\label{tab:noisy_hm3d_val}
	\vspace{-0.6cm}
\end{wraptable}

We evaluated the Context-Waypoint and Context-Trajectory policies (trained with perfect context) in simulation with varying noise levels. For the Context-Waypoint policies, we add shift noise to each of the waypoint inputs. Specifically, at the start of each episode, we obtain waypoints along the shortest path to the goal. We sample noise from a uniform distribution and add the noise to the waypoint. We use the perturbed waypoints as input to the policy. For the Context-Trajectory policy, we add shift noise following the same procedure used for Context-Maps, described in Section \ref{sec:outdated-maps}. We find that these context-variants are more susceptible to noisy inputs, furthering our conjecture that these methods are not suitable to run in the real-world where it's not realistic to have perfect trajectories and waypoints. 
In the presence of noise, the performance of the other context variants drop significantly. We see a maximum decrease of 59.7 SR and 67.8 SPL for Context-Waypoint (row 1 vs 4), and maximum decrease of 49.6 SR, 58.8 SPL for Context-Trajectory (row 5 vs 8). In contrast, Context-Map policies are robust to noise (Table \ref{tab:noisy_hm3d_val}). This points to Context-Maps being the best context modality for sim2real transfer.

\subsection{Improved Robustness for Indoor Navigation using Outdated Maps} 
\label{sec:noisy-maps}

\begin{wraptable}{r}{0.5\textwidth}
	\vspace{-0.6cm}
	\begin{center}
        \resizebox{0.5\textwidth}{!}{
		\begin{tabular}{ccccc}
			\toprule
                \textbf{\#} & \textbf{Train Noise} & \textbf{Eval Noise} & \textbf{SR $\uparrow$} & \textbf{SPL $\uparrow$} \\
			   \midrule
			   1 &  \multirow{4}{*}{20\% Shift} & - & 94.0\tiny{$\pm$1.8} & 73.6\tiny{$\pm$5.8} \\
               2 & & 5\% Shift & 94.0\tiny{$\pm$1.7} & 73.7\tiny{$\pm$5.6} \\
               3 & & 10\% Shift & 93.8\tiny{$\pm$1.6} & 73.9\tiny{$\pm$6.0} \\
               4 & & 20\% Shift & 93.3\tiny{$\pm$1.6} & 73.6\tiny{$\pm$6.2} \\
               \midrule
               5 &  \multirow{4}{*}{50\% Cutout} & - & 91.6\tiny{$\pm$1.3} & 72.4\tiny{$\pm$3.1} \\
               6 & & 10\% Cutout & 91.6\tiny{$\pm$1.6} & 72.7\tiny{$\pm$3.1} \\
               7 & & 25\% Cutout& 92.5\tiny{$\pm$1.1} & 73.1\tiny{$\pm$3.5} \\
               8  & & 50\% Cutout& 92.2\tiny{$\pm$1.1} & 72.6\tiny{$\pm$3.1} \\
			\bottomrule
		\end{tabular}
            }
	\end{center}
	\caption{We evaluate Context-Map policies with varying degrees of noise to the map, and report the average across 3 seeds. Policies trained with noisy maps maintain performance when evaluated with noisy maps.}
	\label{tab:train_noisy_val}
	\vspace{-0.6cm}
\end{wraptable}
In our experiments, we demonstrated that Context-Map policies trained with perfect maps in simulation do not exhibit overfitting. These policies achieve high performance in sim and the real-world when evaluated with inaccurate maps (shown in Table \ref{tab:real_world_res} and \ref{tab:noisy_hm3d_val}). To further improve the robustness of our Context-Map policies, we additionally train Context-Map policies with noisy maps in simulation (using 20\% shift noise, and 50\% cutout noise). We find that these Context-Map policies trained with noise perform just as well as Context-Map trained without noise (Table \ref{tab:noisy_hm3d_val} vs. Table \ref{tab:train_noisy_val}. We see that policies trained with noise are even more robust to varying degrees of noise during test-time (rows 2-4 and 6-8 in Table \ref{tab:train_noisy_val}). In evaluation with varying noise levels, the Context-Map policies trained with noise maintain strong performance (93.3 $\pm$ 1.6 SR at 20\% shift noise, and 92.2 $\pm$ 1.1 SR at 50\% cutout noise).

\subsection{PointNav Policy Architecture}
\label{sec:no-context-arch}
We outline the No-Context PointNav policy architecture from \cite{truong2022kin2dyn}. Given depth observation $d_{t}$, we encode the observation using a 3-layer CNN $\psi$ to obtain depth features $\hat{d}_{t}$.
\begin{equation}
\hat{d}_{t} = \psi(d_t)
\end{equation}
The goal vector $g_{t}$ is represented in polar coordinates $[r, \theta]$, representing the distance and heading relative to the robot's current position. Following \cite{ddppo}, we transform this into $[r, \cos(\theta), \sin(\theta)]$ to account for the discontinuity at the x-axis in polar coordinates. We then pass the goal vector into a full connected layer to get a 32-dimensional output $\hat{g}_{t}$. 
The previous action $a_{t-1}$ is also encoded with a fully connected layer with a 32-dimensional output $\hat{a}_{t-1}$. 

We pass the depth features, goal features, previous action features, and previous hidden state into a 1 layer GRU $\tau^{a}$ with a 512-dimensional hidden size. We denote the hidden state from this GRU $h^{a}_t$.

\begin{equation}
x, h^a_t = \tau(\hat{d}_t, \hat{g}_t, \hat{a}_{t-1}, h_{t-1}).
\end{equation}

The output of the GRU $x$ is fed into two parallel linear layers with a 2-dimensional output size (size of the action space). The outputs $\mu$ and $\sigma$ parameterize a Gaussian action distribution from which the action is sampled. 

\begin{equation}
a_t =  \mathcal{N}(\mu, \sigma) \\
\end{equation}

\subsection{ContextNav Policy Architecture}
\label{sec:context-arch}
We extend the original PointNav policy to incorporate additional environmental context. We use the same architecture for processing the depth obseravtion, goal vector, and previous action as above, and add a separate GRU to process the additional environmental context. We describe the process for encoding 3 types of context input (Context-Map, Context-Trajectory, and Context-Waypoint).

Given a Context-Map or Context-Trajectory of size $2 \times N \times N$ (we use $2 \times 100 \times 100$ as the size of our map), we encode the map using a ResNet18 visual encoder to obtain a 512-dimensional output (context features $\hat{c}_t$). Alternatively, given a Context-Waypoint represented in polar coordinates $[r, \theta]$, we transform this into $[r, \cos(\theta), \sin(\theta)]$. We then pass the waypoint vector into a two-layer MLP to get a 512-dimensional output (context features $\hat{c}_t$). 

We compute the scaled dot-product attention \cite{vaswani2017attention} between the depth and context features to obtain context attention features $\hat{c}^{\text{attn}}_t$ and depth attention features $\hat{d}^{\text{attn}}_t$. 

From a query $Q$ of dimension $d_k$, key $K$, and value $V$, we compute attention using:
\begin{equation}
    \text{Attention}(Q,K,V) = \text{softmax}(\frac{QK^{T}}{\sqrt{d_k}})V
\end{equation}

The query Q is obtained by passing features ($h_t$ or $\hat{c}_t$) through a single linear layer to get $Q_h$, or $Q_c$. The key K is obtained by passing features ($\hat{c}_t$ or $\hat{d}_t$) into a 1D convolution get $K_c$, or $K_d$. The value V is either $\hat{c}_t$ or $\hat{d}_t$.

\begin{equation}
    \hat{c}^{\text{attn}}_t = \text{Attention}(Q_h, K_c, \hat{c}_t)
\end{equation}
\begin{equation}
    \hat{d}^{\text{attn}}_t = \text{Attention}(Q_c, K_d, \hat{d}_t)
\end{equation}

We pass the context and depth attention features, goal features, previous action features, the hidden state from the first GRU ($h^{a}_{t}$), and the hidden state from the previous timestep ($h^{b}_{t-1}$) into a second GRU $\tau^{b}$.

\begin{equation}
x^{b}, h^{b}_t = \tau^{b}(\hat{c}^{\text{attn}}_t, \hat{d}^{\text{attn}}_t, \hat{g}_t, \hat{a}_{t-1}, h^{a}_{t}, h^{b}_{t-1}).
\end{equation}

Following equation 3, the output of the GRU $x^{b}$ is fed into two parallel linear layers whose output parameterizes a Gaussian action distribution from which the action is sampled.

\subsection{Context-Maps in the Real-World}
\label{sec:context-maps-real}
\begin{figure}[h]
  \centering%
  \resizebox{\columnwidth}{!}{
  \renewcommand{\tableTitle}[1]{\huge{#1}}%
  \setlength{\figwidth}{0.3\columnwidth}%
  \setlength{\tabcolsep}{1.5pt}%
  \renewcommand{\arraystretch}{0.8}%
  \renewcommand\cellset{\renewcommand\arraystretch{0.8}%
  \setlength\extrarowheight{0pt}}%
  
    \begin{tabular}{c c c}
  \makecell{\includegraphics[width=0.33\textwidth]{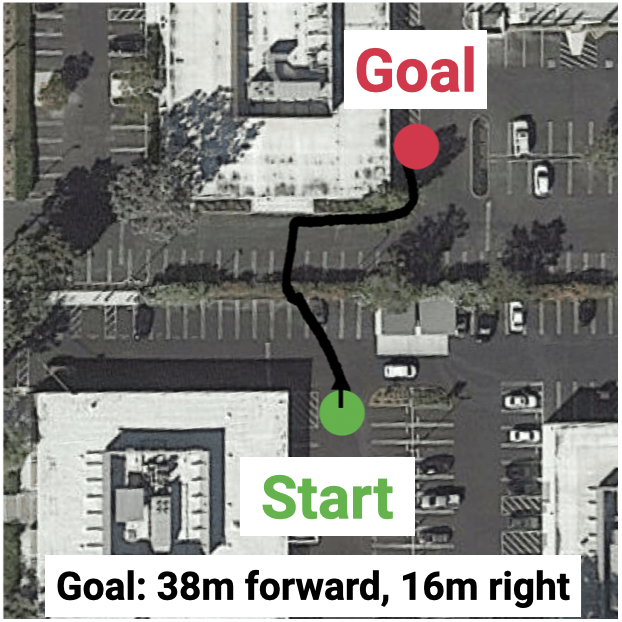}} &
  \makecell{\includegraphics[width=0.33\textwidth]{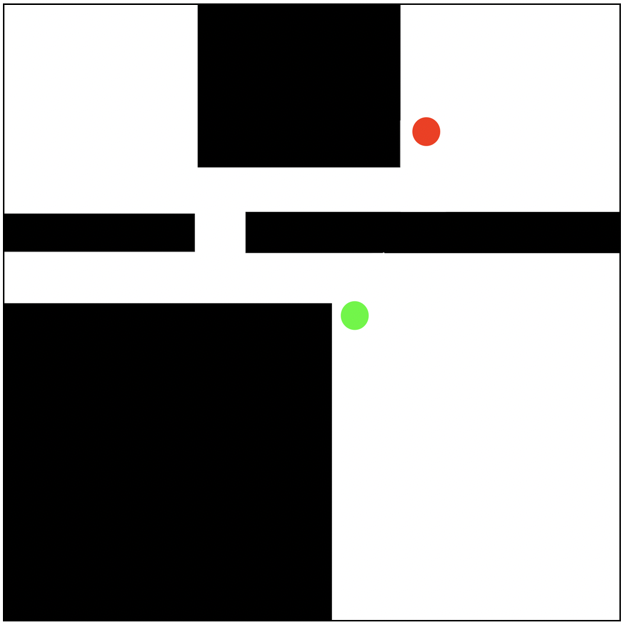}} &
  \makecell{\includegraphics[width=0.33\textwidth]{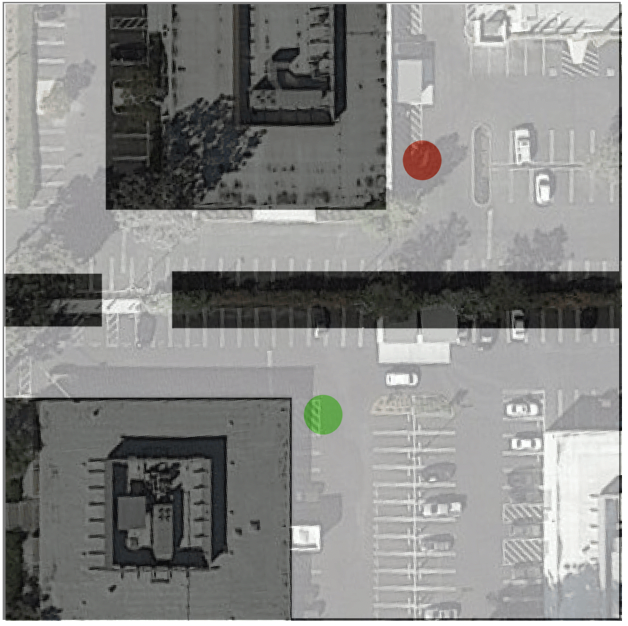}} \\
  \makecell{\includegraphics[width=0.33\textwidth]{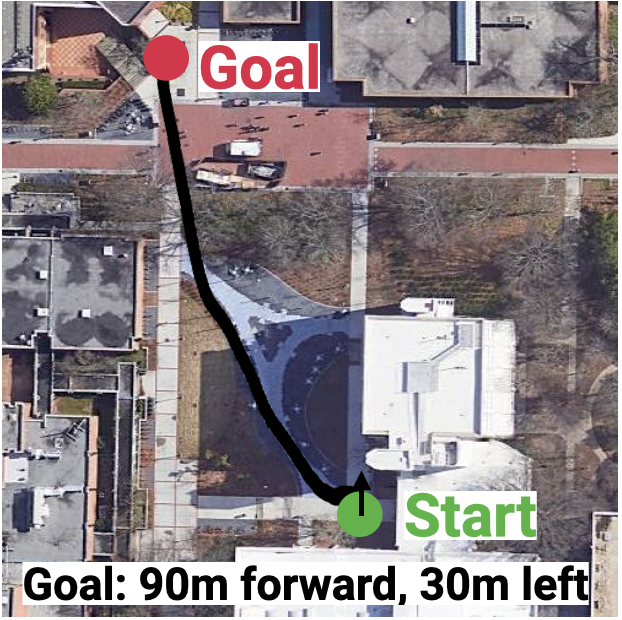}} &
  \makecell{\includegraphics[width=0.33\textwidth]{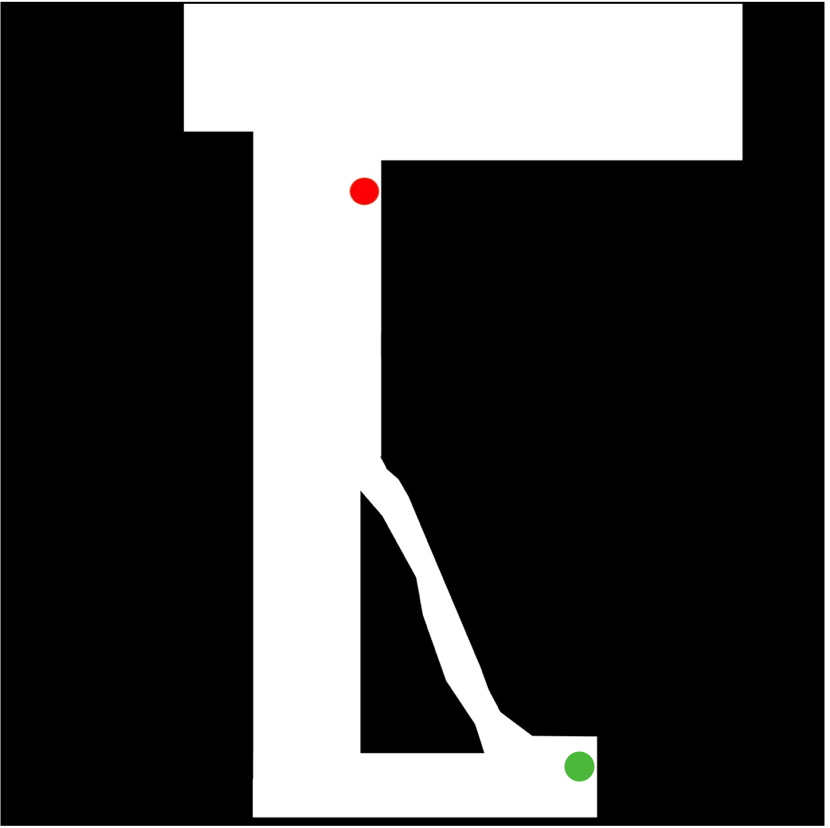}} &
  \makecell{\includegraphics[width=0.33\textwidth]{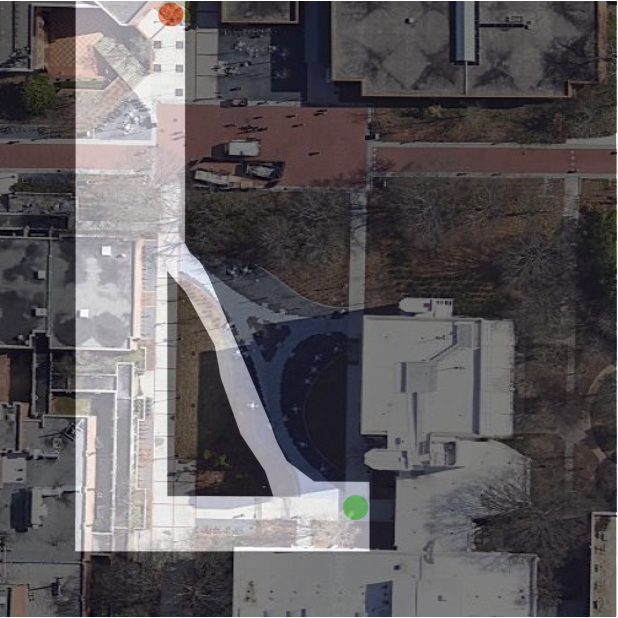}} \\
  \makecell{\includegraphics[width=0.33\textwidth]{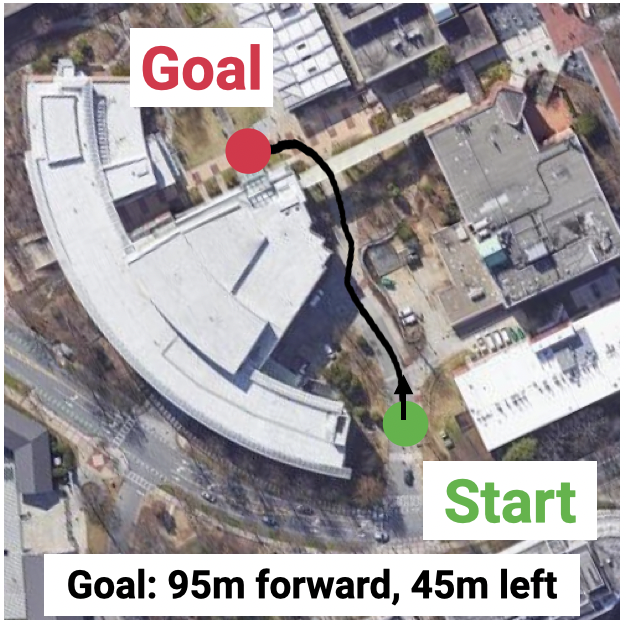}} &
  \makecell{\includegraphics[width=0.33\textwidth]{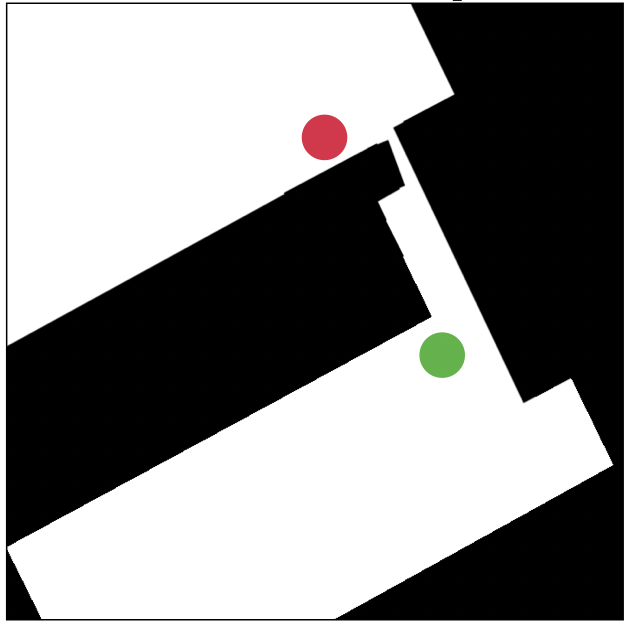}} &
  \makecell{\includegraphics[width=0.33\textwidth]{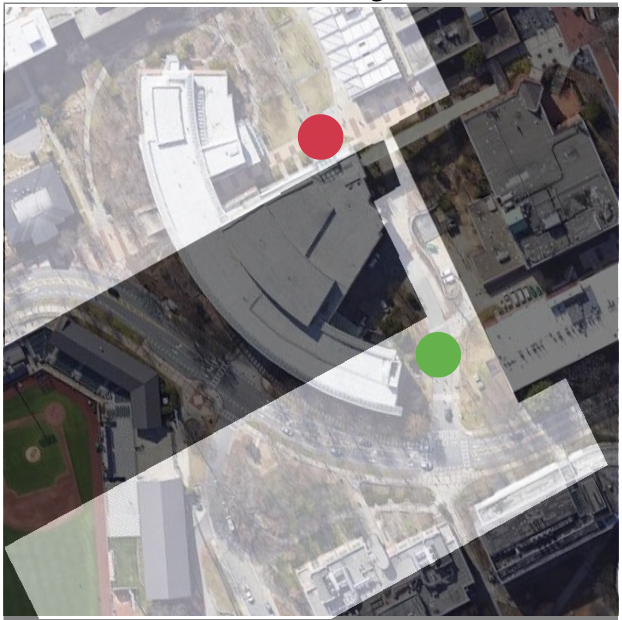}} \\
  Satellite Map & Context-Map & Overlay \\
  \end{tabular}}
  \caption{\textbf{Left:} Satellite images obtained via Google Maps showing the three real-world routes used to test our navigation policies. \textbf{Middle:} We provide outdated Context-Maps to our policy to navigate to the long-range goal. \textbf{Right:} We overlay the Context-Maps on the satellite map to show how outdated, and incomplete the maps are. Our navigation policies can leverage these imperfect maps to successfully navigate to the goal.}
  \label{fig:context_overlays}
\end{figure}

We show the Context-Maps used for each real-world route in Figure \ref{fig:context_overlays}, middle. Starting from the satellite map (first column), a human sketches a map indicating a route for the robot to use to navigate (middle column). We show the context-map overlaid on the satellite map (last column). Notice how inaccurate, incomplete, and outdated the provided context-maps are. The maps hint at a route to the goal for the robot, yet do not show any real-world obstacles that exist on the way to the goal (\eg bushes, cars, tables, chairs, pedestrians).

\begin{figure*}[h]
  \centering%
  \resizebox{\textwidth}{!}{
  \renewcommand{\tableTitle}[1]{\huge{#1}}%
  \renewcommand{\arraystretch}{0.8}%
  \renewcommand\cellset{\renewcommand\arraystretch{0.8}%
  }%
  
    \begin{tabular}{c c c c c c}
   \makecell{\includegraphics[width=0.33\textwidth]{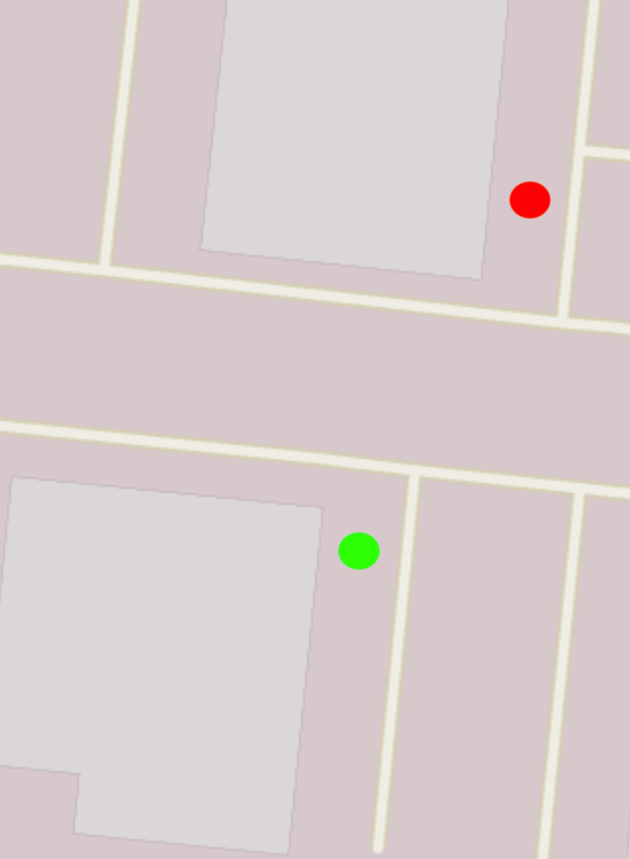}} &
   \makecell{\includegraphics[width=0.33\textwidth]{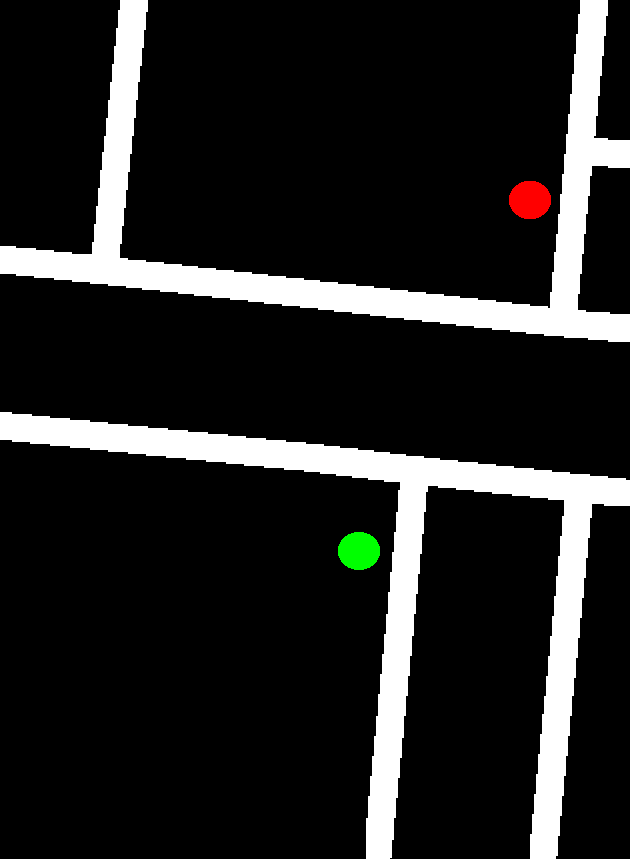}} &
    \makecell{\includegraphics[width=0.33\textwidth]{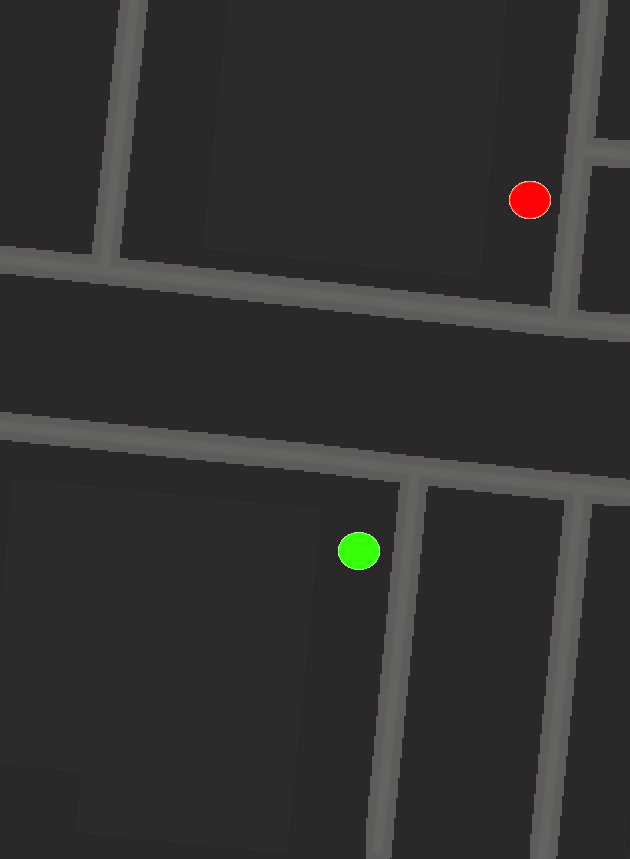}} &
   \makecell{\includegraphics[width=0.33\textwidth]{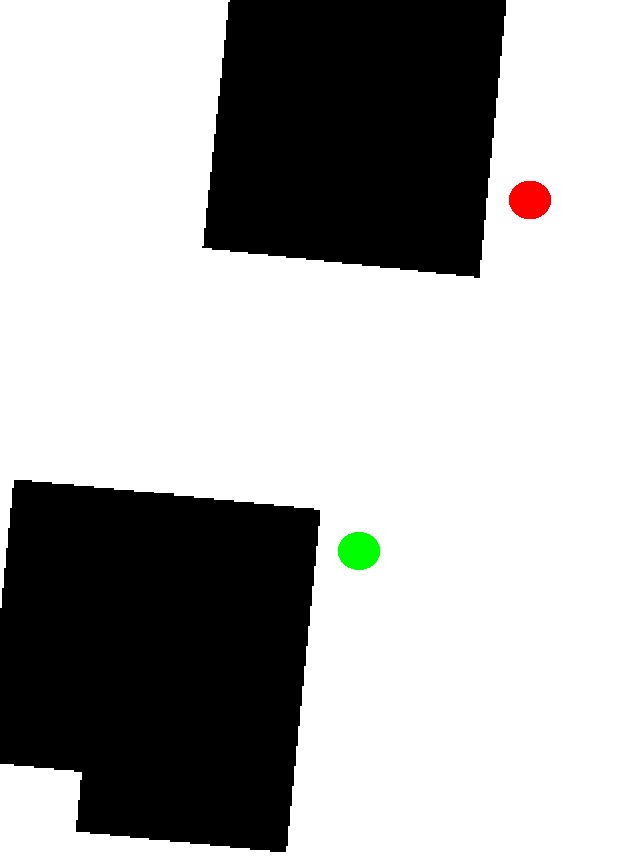}} &
    \makecell{\includegraphics[width=0.33\textwidth]{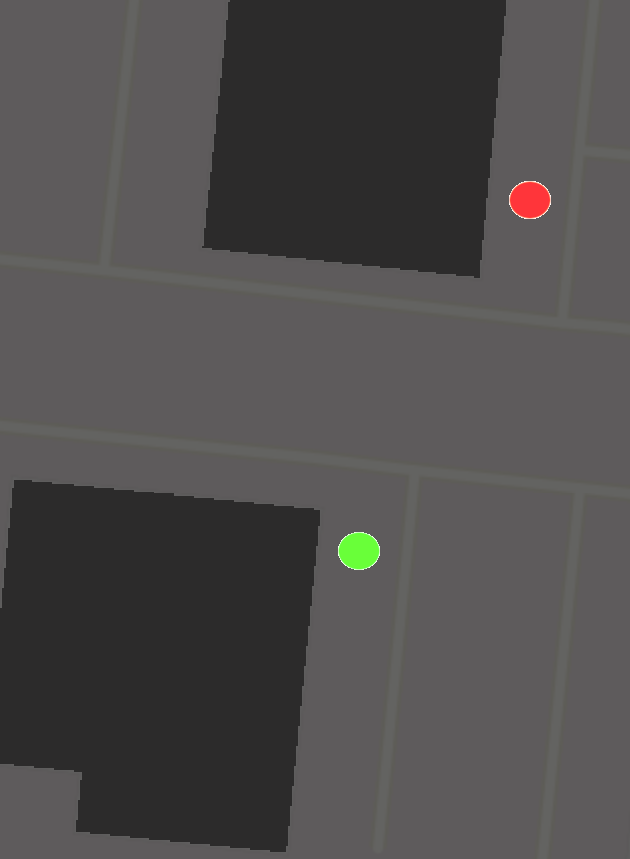}} \\
   \makecell{\includegraphics[width=0.33\textwidth]{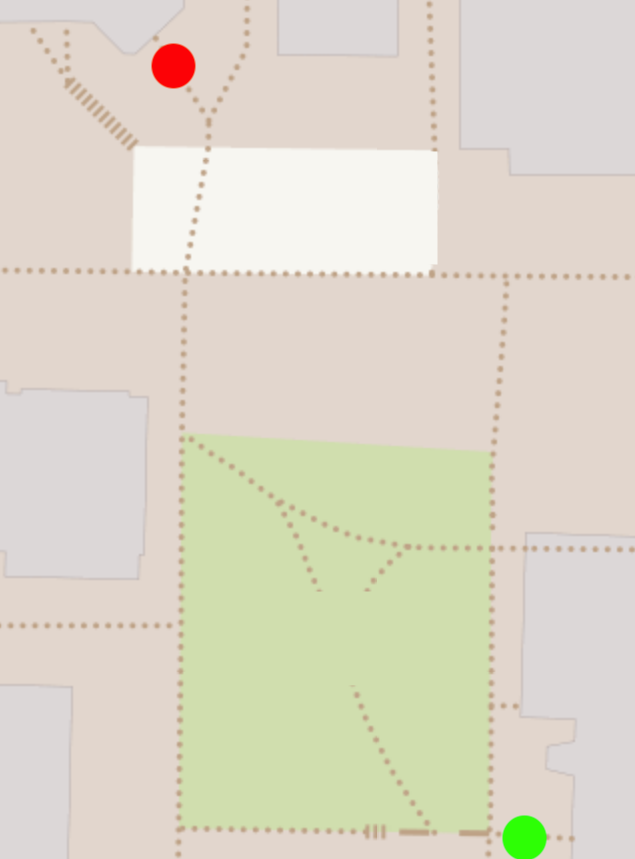}} &
    \makecell{\includegraphics[width=0.33\textwidth]{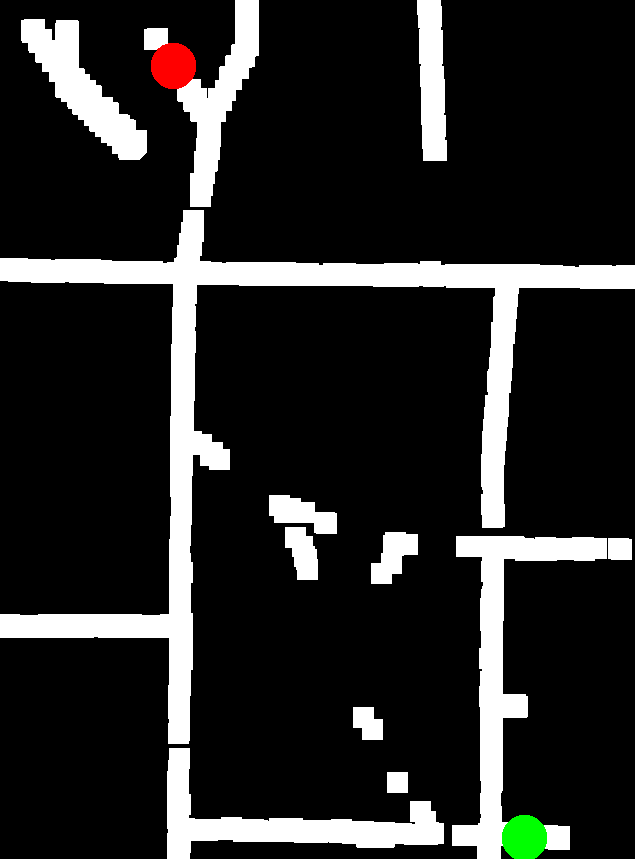}} &
   \makecell{\includegraphics[width=0.33\textwidth]{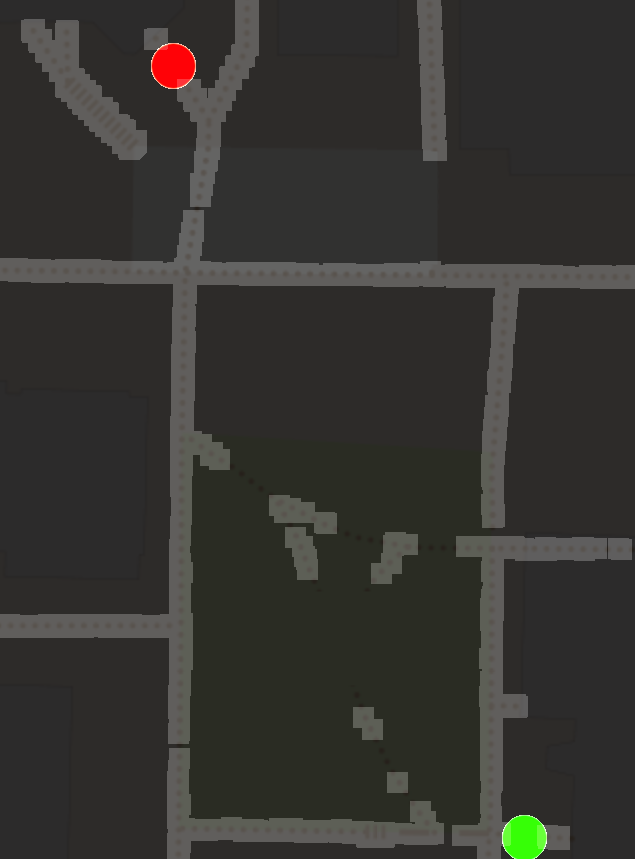}} &
   \makecell{\includegraphics[width=0.33\textwidth]{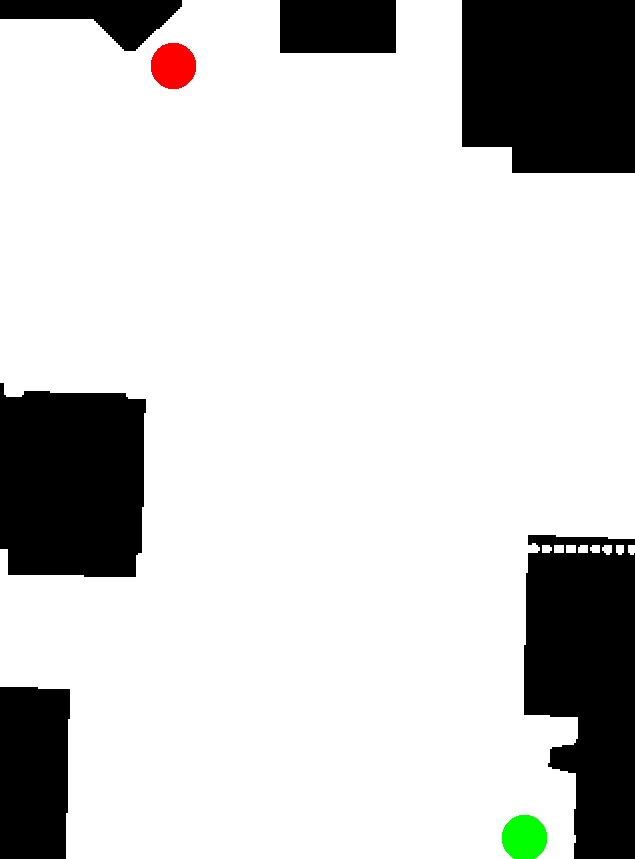}} &
   \makecell{\includegraphics[width=0.33\textwidth]{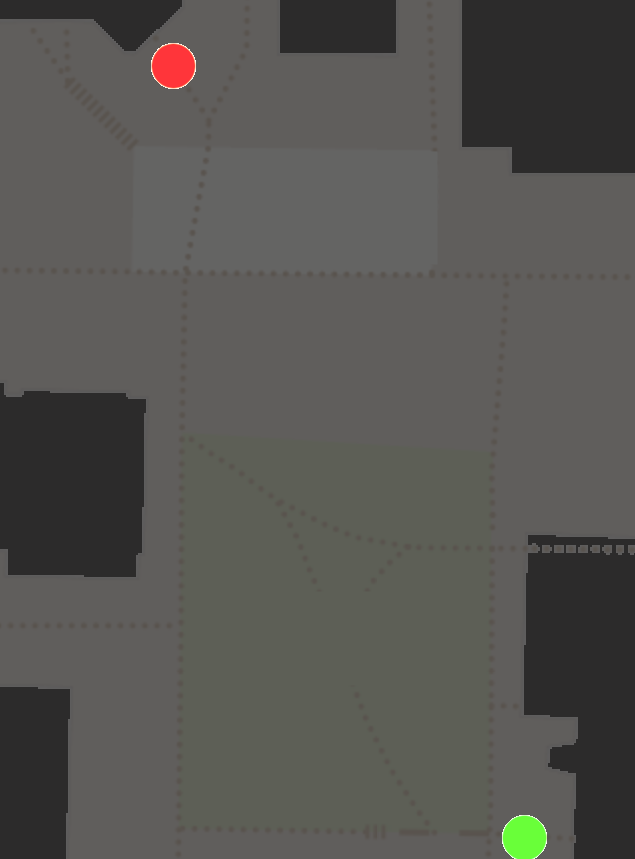}} \\
   \makecell{\includegraphics[width=0.33\textwidth]{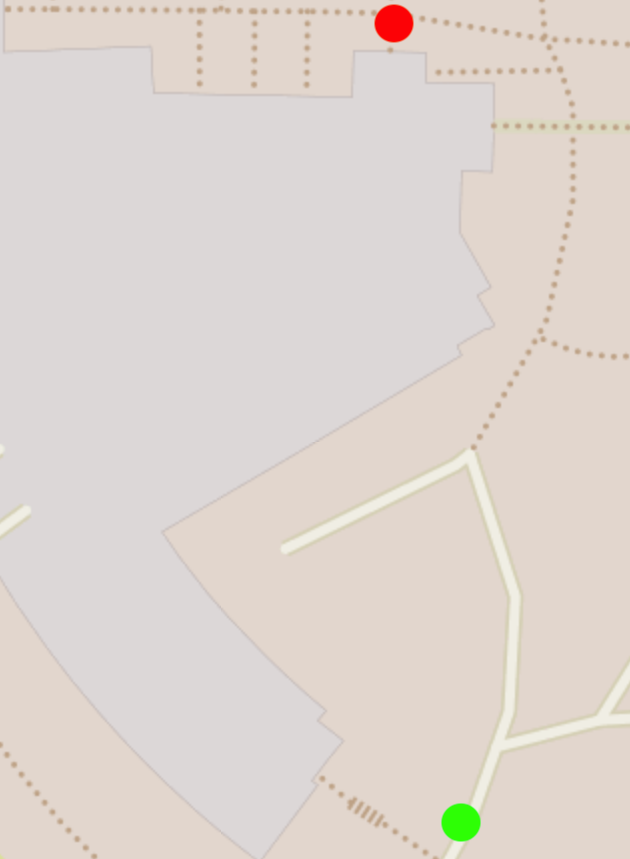}} &
    \makecell{\includegraphics[width=0.33\textwidth]{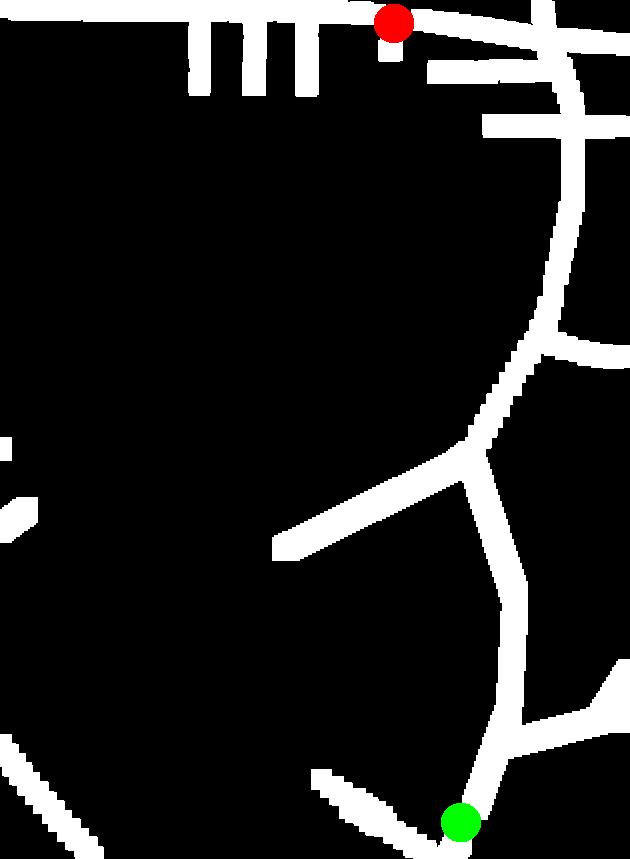}} &
   \makecell{\includegraphics[width=0.33\textwidth]{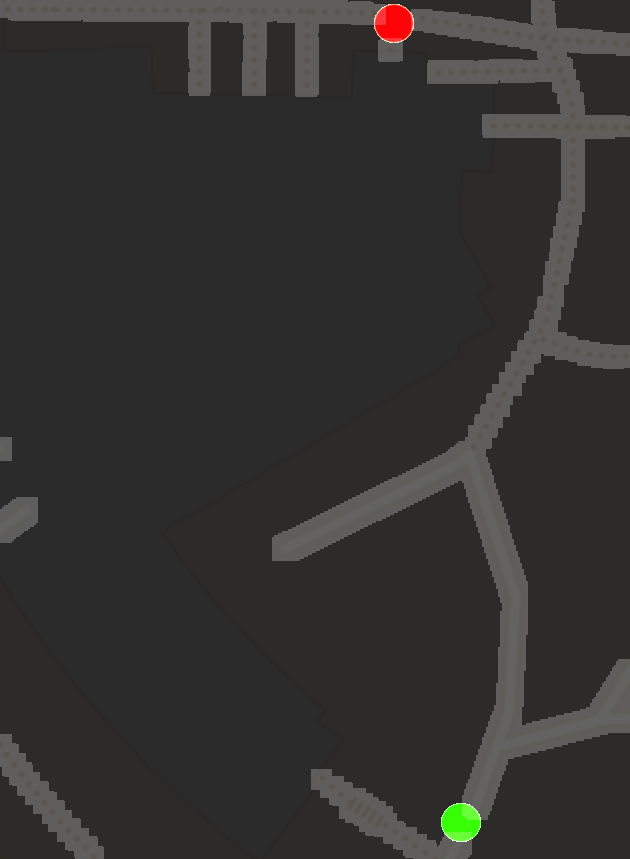}} &
   \makecell{\includegraphics[width=0.33\textwidth]{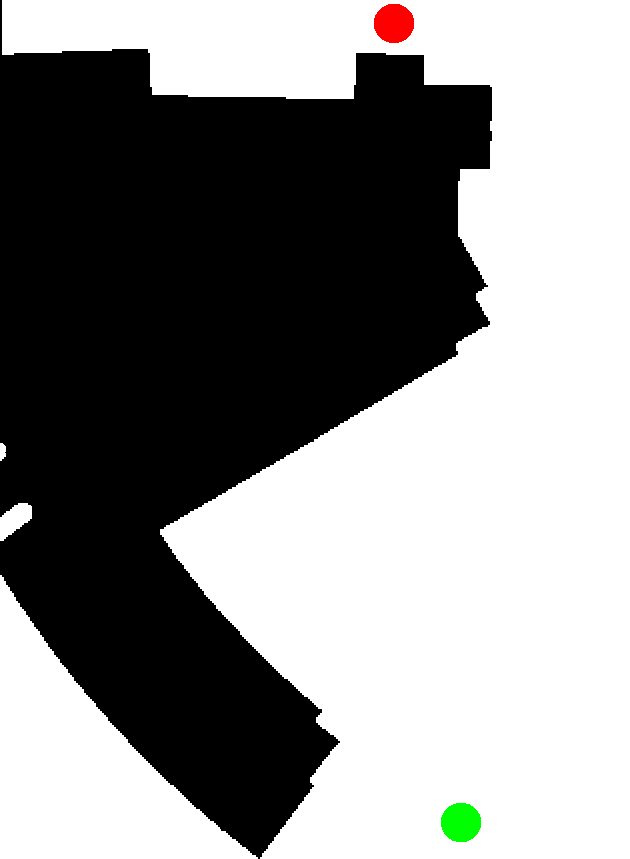}} &
   \makecell{\includegraphics[width=0.33\textwidth]{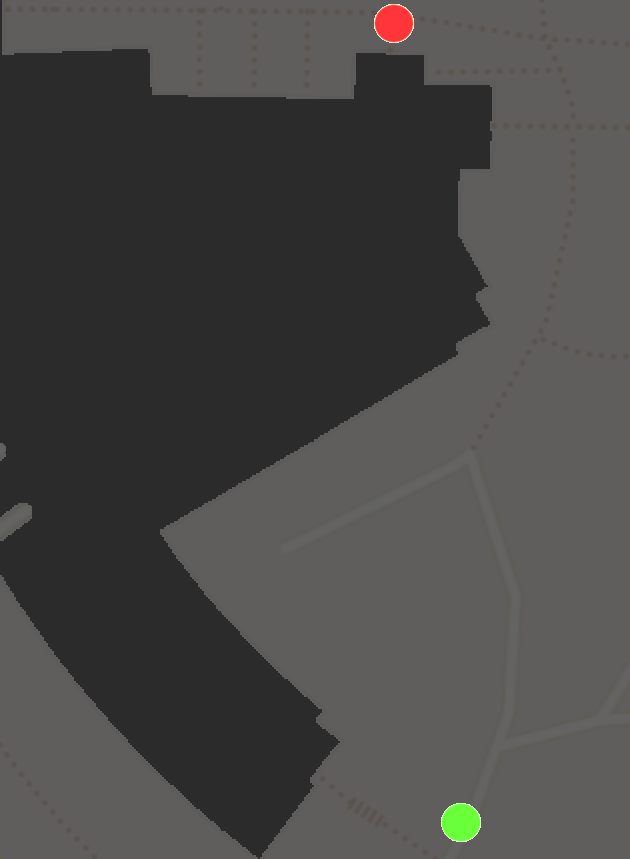}} \\
   \tableTitle{Digital Map} & \tableTitle{Context-Map} & \tableTitle{Overlay} & \tableTitle{Context-Map} & \tableTitle{Overlay} \\
   & \tableTitle{(Routes)} & \tableTitle{(Routes)} & \tableTitle{(Buildings)} & \tableTitle{(Buildings)} \\
  \end{tabular}}
  \caption{We automatically generate Context-Maps for the three routes used in our real-world experiments (routes 1-3 by row). Starting from the digital map (left), we automatically generate context-maps by segmenting out roads (2nd column), or buildings (4th column).}
  \label{fig:auto-maps}
\end{figure*}
\subsection{Auto-Generated Context-Maps}
\label{sec:auto-maps}
We additionally show results of automatically generating context-maps. Starting with digital maps from OpenStreetMap, we can segment out obstacles of interest to obtain a context-map for the navigation route. We show results segmenting out routes (Figure \ref{fig:auto-maps} column 2), or buildings (Figure \ref{fig:auto-maps} column 4). The Context-Map obtained from segmenting routes shows the robot a viable path to the goal, but also contains short segments of other routes in the area that lead to dead ends. The Context-Map obtained from segmenting out buildings gives the robot a high-level idea of the direction to travel to reach the goal. While both variants are useful for navigating on structured paths and large buildings, we note that small man-made openings that are not captured on digital or satellite maps (\ie small opening between bushes that exists for Route 1) are not accurately shown on the map. In this case, the human operator can modify the segmented Context-Map to interject additional paths for the robot to use to navigate to the goal. Using these automatically generated context-maps for robot navigation outdoors is a promising direction for future work. 

\subsection{Boston Dynamics Navigation API}
\label{sec:bd-api}
Boston Dynamics API has two modes for high-level navigation. The first method for navigation is a mapless high-level navigation API (\ie \texttt{go to position}). However, this method cannot navigate around obstacles autonomously, and the robot will stop at the first obstacle it encounters. The second method is Spot's AutoWalk feature in which a user can record and replay navigation missions with Spot. However, this method requires a user to apriori teleoperate Spot through the desired route, and requires fiducials to be placed throughout the desired route to ensure that Spot is properly localized. More fiducuals are needed as the length of the route increases; without the fiducials to localize with, Spot will fail to replay the mission. In the outdoors, different lighting or weather conditions such as strong sunlight or the nightime, may cause the fiducials to not be properly detected by Spot. In contrast, our approach uses zero real-world outdoor experience, and is capable of navigating in different weather conditions. 

\subsection{Reward Function}
\label{sec:reward}
Our reward function is derived from \cite{truong2022kin2dyn}. The reward function is defined as:
\begin{align}
    r_t(a_t,s_t) = R_{success} + R_{geo} + R_{slack} + R_{backward} + R_{coll}
\end{align}
\begin{enumerate}
\item $R_{success}$ is a terminal reward. If the robot successfully completes the episode, it receives a reward of 10.0, otherwise the robot receives 0 reward. \\
\item $R_{geo}$ is a dense reward the robot receives at every step. The robot gets reward based on the change in geodesic distance to the goal from its previous timestep to the current timestep. \\
\item $R_{slack}$ is a slack penalty. This reward encourages the robot to reach the goal using the minimum number of actions. We set the slack penalty to -0.002.\\
\item $R_{backward}$ is a penalty for backwards velocities. We set the backward penalty to -0.3. We found that using this high backwards penalty (10$\times$ higher than the value used in \cite{truong2020learning}) led to higher SPL, as the robot moved backwards less often, which leads to higher path efficiency. \\
\item $R_{coll}$ is a collision penalty. We set the collision penalty to -0.003. We use a small collision penalty in favor of using a higher backwards penalty since collisions were often due to moving backwards. \\
\end{enumerate}
We omit the fall penalty used in \cite{truong2022kin2dyn}. The fall penalty was used for training robots with dynamic control, as the low-level controller may cause the robot to fall over. With kinematic control, the robot is teleported to the next state and would never fall, so the penalty is not needed. 

\subsection{Additional Simulation Details}
\label{sec:sim-details}
\xhdr{SPL Calculation.}
In simulation, we obtain the shortest path in an environment by running A* on the top-down map. This shortest path is used to calculate SPL for our experiments in simulation. We do not report SPL for the outdoor experiments, since we do not have access to the shortest path. Instead, we report the success rate and the distance traveled. 

\xhdr{Context-Maps in Simulation.} 
The context-maps in simulation are binary top-down maps that were generated by sampling a navigation mesh with a vertical slack. We use 0.5m, the default in Habitat.

\end{document}